\documentclass[letterpaper, 10 pt, conference]{ieeeconf}
\usepackage{times}
\usepackage{helvet}
\usepackage{courier}
\usepackage{color}
\usepackage{amsmath}
\usepackage{amssymb}
\usepackage[ruled,vlined,linesnumbered]{algorithms2e}
\usepackage{graphicx}

\newtheorem{theorem}{Theorem}

\long\def\omitit#1{}

\begin{document}

\title{A Graph Isomorphism-based Decentralized Algorithm for Modular Robot Configuration Formation}

\author{Ayan Dutta$^{1}$, Prithviraj Dasgupta$^{1}$ and Carl Nelson$^{2}$\\
$^1$ Computer Science Department, University of Nebraska at Omaha\\
             $^2$ Mechanical and Materials Engineering 
  Department, University of Nebraska-Lincoln.\\
  Email: \{adutta, pdasgupta\}@unomaha.edu, cnelson5@unl.edu
}

\maketitle

\begin{abstract}
We consider the problem of configuration formation in modular robot systems where a set of modules that are initially in different configurations and located at different locations are required to assume appropriate positions so that they can get into a new, user-specified, target configuration. We propose a novel algorithm based on graph isomorphism, where the modules select locations or spots in the target configuration using a utility-based framework, while retaining their original configuration to the greatest extent possible, to reduce the time and energy required by the modules to assume the target configuration. We have shown analytically that our proposed algorithm is complete and guarantees a Pareto-optimal allocation. Experimental simulations of our algorithm with different number of modules in different initial configurations and located initially at different locations, show that the planning time of our algorithm is nominal (order of msec for $100$ modules). We have also compared our algorithm against a market-based allocation algorithm and shown that our proposed algorithm performs better in terms of time and number of messages exchanged.
\end{abstract}


\section{Introduction}
\begin{figure}[t!]
\begin{tabular}{cc}
\hspace{-0.2in}
\includegraphics[width=1.75in]{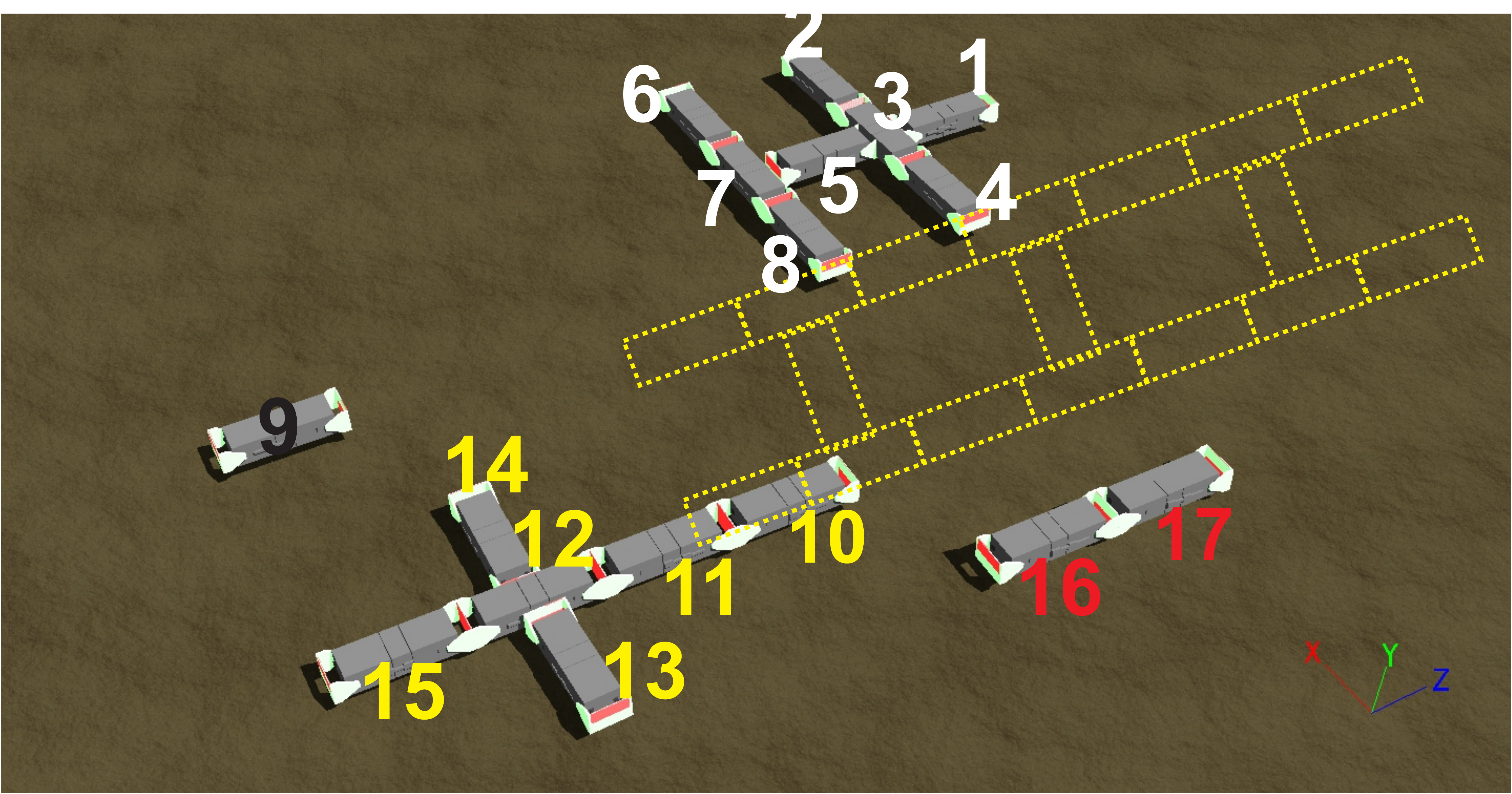}&
\hspace{-0.1in}
\includegraphics[width=1.75in]{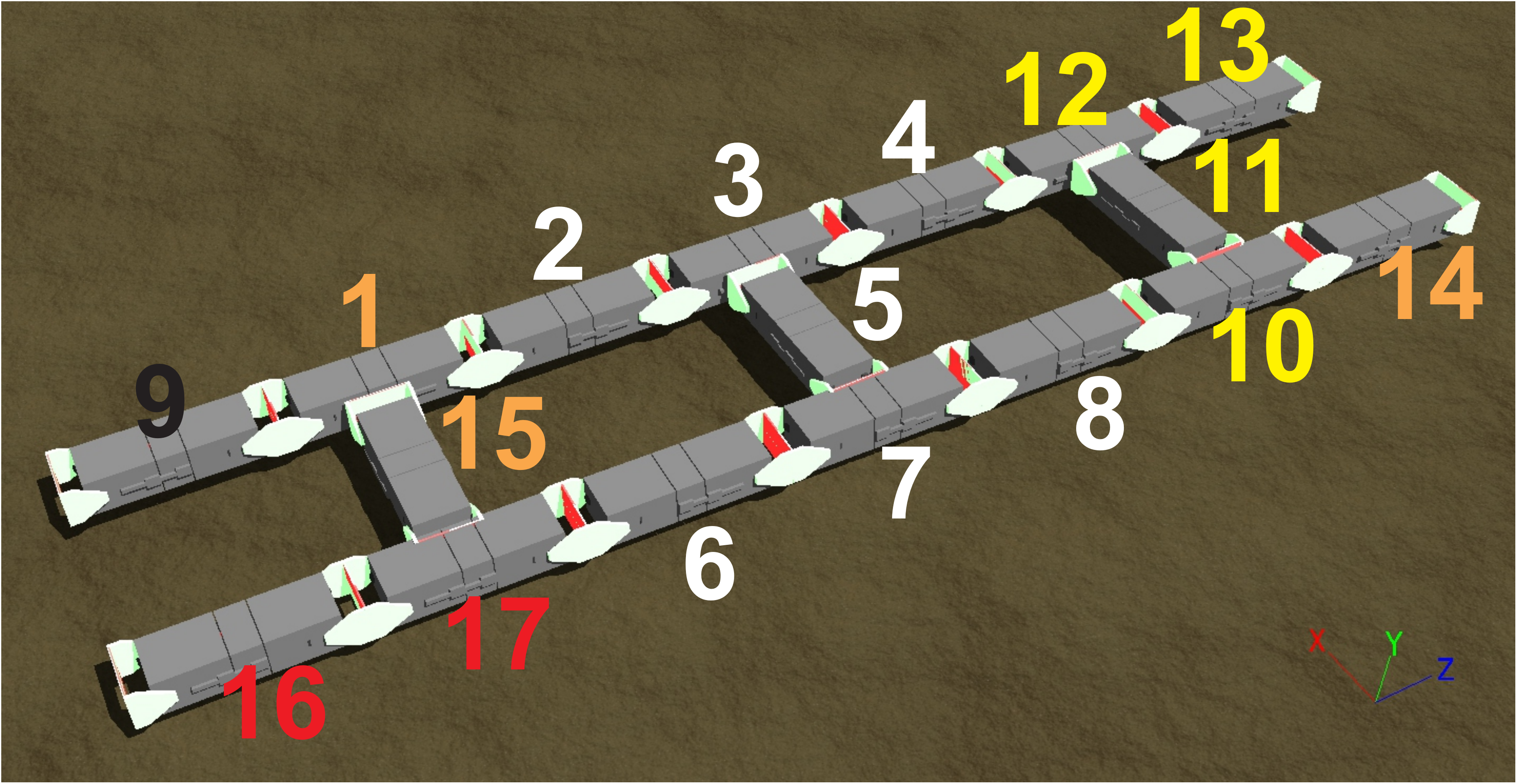}\\
{\small{(a)}} & {\small{(b)}} 
\end{tabular}
\caption{{\small{(a) Four initial configurations consisting of $1$, $2$, $6$ and $8$ modules respectively, and desired target configuration (marked with yellow dotted lines) (b) target configuration involving all $17$ modules connected in ladder configuration; module numbers marked in white, yellow and red are retained between between initial and target configurations.}}}
\label{fig_example}
\end{figure}

Modular self-reconfigurable robots (MSRs) \cite{Yim07} are composed of individual robotic modules which can change their connections with each other to form different shapes or configurations. This configuration adaptability affords a high degree of dexterity and maneuverability to MSRs and makes them suitable for robotic applications such as inspection of engineering structures like pipelines \cite{Choset13}, extra-terrestrial surface exploration \cite{Chien05}, etc. A central problem in MSRs is to autonomously reconfigure the modules from one configuration to another~\cite{Stoy10}. In this paper, we consider an aspect of this MSR reconfiguration problem called the configuration formation problem, which is described as follows: we are given a set of modules forming different configurations that are distributed at different locations within the environment along with a target configuration that needs to be formed at a specified location; the target configuration involves some or all of the modules from the initial configurations. The problem is to select an appropriate subset of modules to occupy appropriate spots or positions in the target configuration, so that, after reaching the selected positions, they can readily connect with adjacent modules and form the shape of the desired target configuration. An instance of this problem is shown in Figure \ref{fig_example}.

Previous research in MSR configuration formation has mainly focused on the problem of {\em self}-reconfiguration, where the objective is to transform one configuration to another without removing or adding modules to the configuration; only the positions of the modules in the initial configuration are changed to obtain the target configuration. Our work in this paper generalizes the configuration formation problem to a scenario where the number of modules in the target configuration is independent of the number of modules in the initial configuration, and, individual modules or a connected set of modules can be extracted from multiple initial configurations to form the target configuration. The generalized configuration formation problem is non-trivial as the modules might already be connected in initial configurations that do not correspond to parts of the target configuration. Also, existing connections between modules in the initial configuration should be preserved in the target configuration, whenever possible, to reduce the energy and time expenditure in disconnecting and re-connecting modules. Moreover, multiple modules from different initial configurations might end up selecting the same most-preferred position (e.g., position involving least time and battery expenditure to navigate to) in the target configuration, leading to failed attempts to achieve the target configuration. To address these challenges, we propose an algorithm that allows modules from initial configurations to select suitable positions in a target configuration using a technique based on graph isomorphism that attempts to improve the utility of the modules by reducing the number of disconnects between modules to achieve the target configuration. We have shown analytically that our proposed algorithm is complete and achieves Pareto-optimal allocation. We have also verified the performance of our algorithm in terms of planning time and number of messages exchanged for different number of modules and different initial and target configurations, for simulated modules of the MSR. 
Our experimental results show that our algorithm performs better, in terms of planning time and number of messages passed by the modules as compared to a market-based allocation algorithm.


\section{Related Work}
An excellent overview of the state of the art MSRs and their self-reconfiguration techniques is given in~\cite{Stoy10}. One of the popular techniques for MSR self-reconfiguration is AI-based search techniques such as Rapidly-exploring Random Trees (RRT) and A*~\cite{Brandt06}. However, as noted in a recent survey on MSRs~\cite{ahmadzadeh2015modular}, configuration formation in modular robot systems has been studied less extensively. Alonso-Mora {\em et al.}~\cite{alonso2011multi} addressed a position selection problem for artistic pattern formation by a swarm of robots where goal positions for robots are specified as Voronoi regions and the Hungarian algorithm~\cite{hungarian} is used to allocate robots to goal positions. In~\cite{werfel2008three}, the authors have provided decentralized movement strategies for robots using random walk, systematic search, or, gradient-following to enable them to carry blocks to build user-specified configurations. A potential limitation of these approaches when applied to MSR configuration formation is that it would require modules already connected in a certain initial configuration to be first disconnected into singletons and then allocated to individual spots in the target configuration, resulting in unnecessary expenditure of energy to undock modules in the initial configuration and possibly re-dock the same modules in the target configuration; inter-module collision avoidance during locomotion of multiple individual modules would also consume more time and energy than when the same modules move together as a connected configuration. In contrast, our proposed approach attempts to preserve initial configurations in parts of the target configuration wherever possible using graph isomorphism~\cite{cordella2004sub} to avoid these issues. Graph isomorphism for MSRs has been investigated by several researchers~\cite{Nelson04,Park08,Hou14}, albeit for self-reconfiguring modules (changing positions of modules) that remain part of the same configuration after reconfiguration. Our proposed approach generalizes this direction of research by finding the best positions for multiple configurations and singleton modules within a different, possibly larger or smaller, target configuration.

\section{Configuration Formation as Utility Maximization Problem}
\label{sec_model}
Let $\mathbb{A}=\{a_1, a_2,...\}$ denote a set of robot modules. Each $a_i \in \mathbb{A}$ has an initial pose denoted by $a_i^{pos} = (x_i, y_i, \theta_i)$, where $(x_i, y_i)$ denotes the location of $a_i$ and $\theta_i$ denotes its orientation within a 2-D plane corresponding to the environment. Each module has a unique identifier. A configuration is a set of modules that are physically connected. A configuration is denoted as $A_i = \{a_1, a_2,..,a_j\} \subseteq \mathbb{A}$. The topology of configuration $A_i$ is denoted as a graph, $G_{A_i}=(V_A,E_A)$, where $V_A= A_i$ and $E_A=\{e_{kj}=(a_k, a_j):\;\mbox{if}\; a_j\,\mbox{and}\,a_k\, \mbox{are connected in}\,A_i\}$. Each configuration has a module that is identified as a leader~\cite{baca2014} and the leader's pose is used to represent the configuration's pose.

In the configuration generation problem studied in this paper, robot modules, starting from a set of different initial configurations, are required to get into a specified target configuration. The target configuration is also represented as a graph, denoted by $G_T=(V_T,E_T)$, where $V_T= \{s_1, s_2,...\}$ is the set of vertices and $E_T=\{e_{ij}=(s_i, s_j)\}$ is the set of edges. Each vertex in $V_T$ is referred to as a {\em spot} that a module needs to occupy and two neighboring spots share an edge between them depending on the topology of $G_T$. Each spot $s_i \in V_T$ is specified by its pose and its neighboring spots in the target configuration, $s_i = (s_i^{pos}, neigh(s_i))$, where $neigh(s_i) \subset V_T$. In the rest of the paper, for the sake of legibility, we have slightly abused the notation by using $T$ instead of $G_T$ to denote the target configuration and $S$ instead of $V_T$ to denote the spots in the target configuration.

To formulate the configuration generation problem as a utility maximization problem, we first represent the utility of a single module to occupy a single spot in the target configuration, and, then extend that representation to a set of modules connected as a configuration to occupy a set of adjacent spots in the target configuration. A single module's utility for a spot is given by the value of the spot to the module minus the costs or energy expended by the module to occupy the spot. As reported in~\cite{kamimura2004distributed}, the locomotion of an MSR is significantly affected by the locomotion of the module(s) in the MSR that has more neighbors in the MSR's configuration. For example, for the target configuration shown in Figure \ref{fig_example}(a), the module $12$ at the center of the $6$-module configuration is more critical than the other modules for locomotion as it has more neighbors. To capture this position dependency, we have used a concept from graph theory called the betweenness centrality~\cite{brandes2001faster} to denote the value of spot $s_i$, given by: $Val(s_i) = \frac{\sigma_{s_j\,s_k}(s_i)}{\sum \limits_{s_i \not= s_j \not= s_k} \sigma_{s_j\,s_k}},$ where $\sigma_{s_j\,s_k}$ is the total number of shortest paths between any pair of nodes $s_j$ and $s_k$ in $G_T$ and $\sigma_{s_j\,s_k}(s_i)$ is the number of shortest paths between $s_j$ and $s_k$ which go through $s_i$. The cost to a module $a_i$ located at $a_i^{pos}$ to occupy spot $s_j$ at $s_j^{pos}$, is calculated as a sum of $a_i$'s locomotion costs to reach and occupy spot $s_j$, and any costs to undock and re-dock with neighboring modules before and after it occupies the spot~\cite{Dasgupta12}. This is denoted as $cost_{a_i}(s_j) = cost^{loc}(a_i^{pos}, s_j^{pos})+ \sum \limits_{a_k \in neigh(s_j)} cost^{dock}(a_i, a_k)+\sum \limits_{a_{i'} \in neigh(a_i)} cost^{undock}(a_i, a_{i'})$, where $cost^{loc}()$ denotes the locomotion cost from $a_i^{pos}$ to $s_j^{pos}$, $cost^{dock}$ denotes the cost of docking $a_i$ with modules in neighboring spots of $s_j$ and $cost^{undock}$ denotes the undocking costs of $a_i$ from neighboring modules in $A_i$. Note that energy requirements for locomotion of a module are generally higher than those for docking the module with another module as locomotion requires continuous power to all motors and much higher torques than docking; also, docking two modules requires aligning their docking ports first, which takes more energy than un-docking two modules. Therefore, $cost^{loc}() \gg cost^{dock}() > cost^{undock}()$.

When a set of modules is connected in configuration $A_i$, the cost of occupying a set of spots $S_j \subseteq V_T$ in the target configuration is given by: $cost_{A_i}(S_j) = \sum \limits_{s_l \in S_j, a_k \in A_i} cost_{a_k}(s_l) - f_{rwd}(|A_i|)$, where $f_{rwd}(|A_i|) = \frac{|A_i|-2}{|\mathbb{A}|}$ is a reward function for retaining connections between modules in the existing configuration $A_i$ into smaller configurations. Because $f_{rwd}(|A_i|)$ increases (and $cost_{A_i}()$ decreases) with the size of $A_i$, it is cost-wise better to break smaller configurations than to break larger configurations to fit into the target configuration. So, the reward function ensures that keeping the initial configuration intact in the target configuration, whenever possible, results in lower cost. Using the above formulation, it can easily be seen that when $A_i$ can fit entirely into $V_T$ (i.e., $S_j = V_T$), $cost_{A_i}(S_j) < \sum \limits_{s_j \in S_j, a_i \in A_i} cost_{a_i}(s_j)$. 

The utility of a spot to a module determines how profitable or beneficial that spot is for the module if it finally ends up occupying that spot. The utility of module $a_i$ for spot $s_j$ is given by $U_{a_i}(s_j) = Val(s_j) - cost_{a_i}(s_j)$. Similar to the cost function described above, the utility for initial configuration $A_i$ to occupy a set of spots $S_j \subseteq V_T$, is given by the sum of the utilities of the individual modules comprising $A_i$ to occupy spots in $S_j$, $U_{A_i}(S_j) = \sum \limits_{s_l \in S_j} Val(s_l) - cost_{A_i}(S_j)$. Using the above formulation, the spot allocation problem has to assign modules to spots so that each module is allocated to the most eligible (highest utility earning) spot and no two modules are assigned to the same spot. Given a set of modules $\mathbb{A}$ in a set of initial configurations, and, a set of spots $S$ representing the target configuration, find a suitable allocation $P^* : \mathbb{A} \rightarrow S$ such that $P^* = \displaystyle \arg \max_{\forall P} (\sum_{a_i \in \mathbb{A}, s_j \in S} U_{a_i}(s_j) + \sum_{A_i \subseteq {\mathbb{A}}, S_j \subseteq S} U_{A_i}(S_j)$ and $\forall a_k \neq a_i, \quad P^*(a_i) \neq P^*(a_k)$. Note that, if two modules $a_i$ and $a_k$ both have the same highest utility for spot $s_j$, then only one of them can be allocated to and occupy $s_j$. In the next section, we describe our spot selection algorithm that provides a suitable allocation of modules to spots for the above utility maximization problem. 


\section{Spot Selection Algorithm}

We divide the problem into two phases - a {\it planning phase}, where modules select spots in the target configuration, and an {\it acting phase}, where modules move to their selected spots and connect with other modules.

\subsection{Planning Phase}
In the beginning of the planning phase, all the modules broadcast their positions and orientations. After having this information, each module calculates the location corresponding to the center target configuration $T$ in the environment, as the mean of all spots' positions.\footnote{A common coordinate system can be maintained by modules for localizing themselves following the model described in \cite{suzuki1996agreement}.} Individual modules then rank themselves according to their distances from the center of $T$; the rank of a configuration is calculated using the distance of the configuration's leader from the center of $T$. Modules and configurations select spots in $T$ based on their rank. Because $cost^{loc}$ has the most significant contribution to the cost function, the distance-based rank ensures that modules and configurations with lower costs (higher utilities) get to select spots in $T$ first. We describe the spot selection techniques in the planning phase in two parts - spot selection by singleton modules and spot selection by configurations. 

\begin{algorithm}[ht!]
\textbf{procedure:} {\em spotAllocation()}\\
\KwIn{$S$: set of spots, $\bar{S}$: set of (spot, selector) pairs; $a_{curr}$: robot currently selecting spot.}
$S_{sort} \leftarrow$ Sort $S$ in descending order of utility of spots\\
\For{each $s_j \in S_{sort}$}{
$\mathcal{D} \leftarrow 0$;\\
\If{($s_j$ is not selected by another module) $\vee$ (($s_j$ is selected by module $a_{block} \notin A_i \subseteq \mathbb{A}) \wedge$ (evict($a_{curr}, a_{block}$, $\mathcal{D}$) = TRUE))}
	{
		Select spot $s_j$ for $a_{curr}$;\\
		Broadcast updated set of spot-selector pairs $\bar{S}$;\\
		return;
	}
}
Broadcast NO\_SPOT\_FOUND message;
\newline
\newline
\textbf{procedure:} {\em evict($a_{curr},a_{block},\mathcal{D}$)}\\
\KwIn{$a_{curr}$: robot currently selecting spot $s_{curr}$; $a_{block}$: the robot which has already selected $a_{curr}$'s best spot $s_{curr}$; $\mathcal{D}$: current recursion depth.}
\If{ $\mathcal{D} < \mathcal{D}_{max}$}{
	$s_{block} \leftarrow$ arg $\max \limits_{s_i \in S \setminus s_{curr}} U_{a_{block}}(s_i)$;\\
	$s_{curr'} \leftarrow$ arg $\max \limits_{s_i \in S \setminus s_{curr}} U_{a_{curr}}(s_i)$;\\
	\If{($U_{a_{curr}}(s_{curr}) + U_{a_{block}}(s_{block}) > U_{a_{curr}}(s_{curr'}) + U_{a_{block}}(s_{curr}))$}{
		\eIf {$s_{block}$ is not selected by any module}
		{return TRUE;}
		{
		//$a_{block}' \notin A_i \subseteq \mathbb{A}$ is the module occupying $s_{block}$\\
		return evict($a_{block}, a_{block}',\mathcal{D}+1$);
		}
	}
}
return FALSE;

\caption{Spot Allocation Algorithm for Singleton Modules and Eviction algorithm used by modules to select alternate spots}
\label{algo_singleton}
\end{algorithm}

\subsubsection{Spot Selection by Singleton Modules}
A singleton module $a_{curr}$ selects a spot to occupy using Algorithm \ref{algo_singleton}. $a_{curr}$ first sorts the spots in order of its expected utility $U_{a_{curr}}(s_j), \forall s_j \in S$. If a spot $s_j$ has not already been selected by another module, or, if it has been selected by another singleton module (module that is not part of a configuration) that can be evicted using the {\em evict} method, then $a_{curr}$ selects $s_j$ and broadcasts the updated spot-selector pairs to all other modules. If $a_{curr}$ cannot evict the module currently occupying its highest utility spot, then it successively reattempts spot selection using the spots for which it has the next highest utilities. If none of the spots in $S$ can be selected by $a_{curr}$, it broadcasts a NO\_SPOT\_FOUND message to all other modules.

\begin{figure}[ht!]
\begin{center}
\hspace{-0.2in}
\includegraphics[width=3.5in]{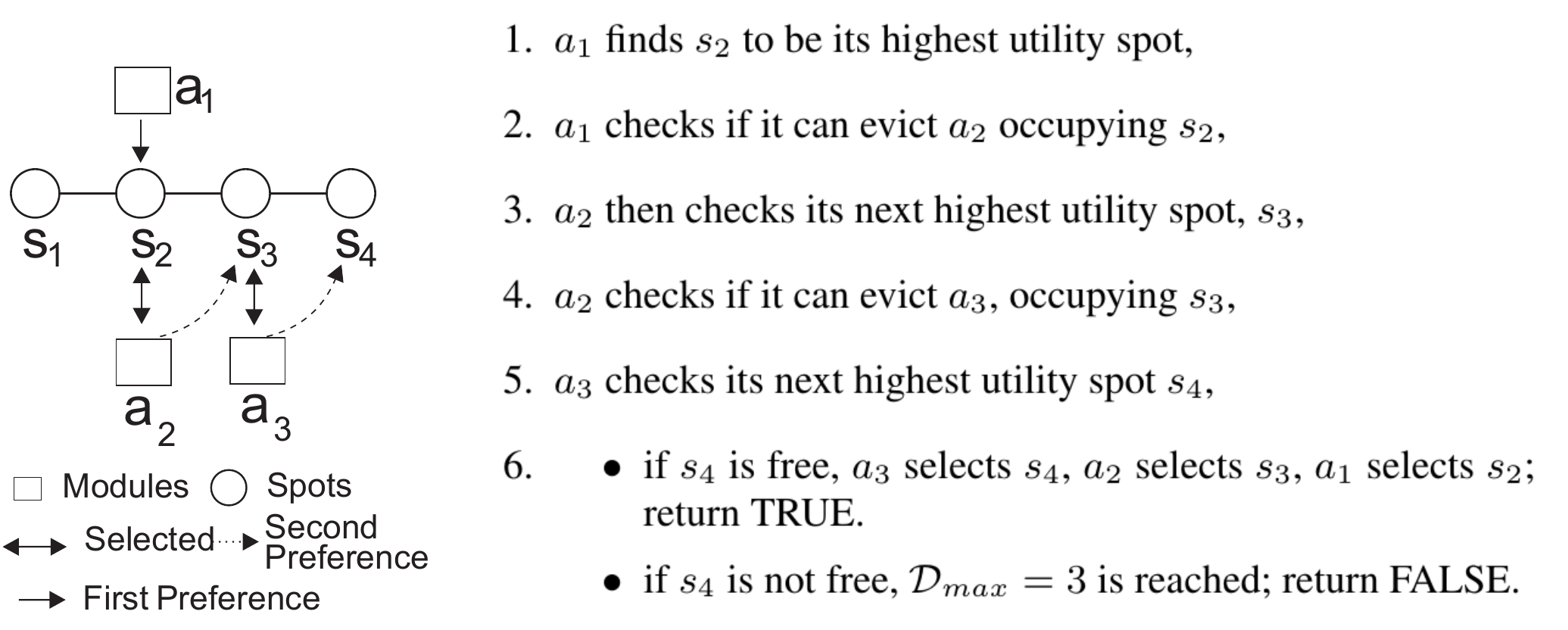}
\end{center}
\caption{Illustration of eviction algorithm for $3$ modules with $\mathcal{D}_{max}=3$.}
\label{eviction_fig}
\end{figure}

The {\em evict method} is used by module $a_{curr}$ to cancel the selection of spot $s_{curr}$ done previously by another singleton module $a_{block}$. Note that eviction can be done only for a singleton module, and not for modules that are part of configurations, as breaking existing configurations will incur additional time as well as costs for docking and undocking modules. The method first checks the expected combined utility between $a_{curr}$ and $a_{block}$ for selecting their most (conflicting) and second-most preferred spots. If this combined utility is greater when $a_{curr}$ selects $s_{curr}$ and  $a_{block}$ selects its next highest utility spot that it can occupy, then $a_{curr}$ evicts the selection of $s_{curr}$ by $a_{block}$, as shown in the {\em evict()} method in Algorithm \ref{algo_singleton}. To limit excessively long cycles of eviction, we have allowed at most $\mathcal{D}_{max}$ successive evictions. An illustration of the eviction process with $\mathcal{D}_{max}=3$ is shown in Figure \ref{eviction_fig}. 

\subsubsection{Block Allocation by Modules Connected in a Configuration}
The technique used by configuration $A_{curr}$ to select a set of connected spots in the target configuration $T$ is given by the {\em blockAllocation} algorithm shown in Algorithm \ref{algo_configuration}. The algorithm is executed on $l_{curr}$, the leader of configuration $A_{curr}$, selected using techniques in ~\cite{baca2014}.

\begin{figure}
\begin{center}
\begin{tabular}{ccc}
\includegraphics[width=0.3\linewidth]{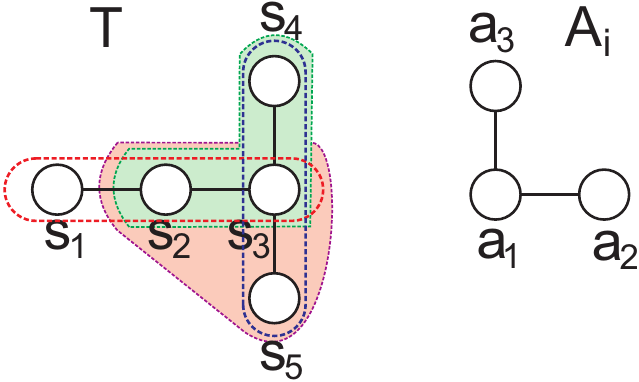}&
\includegraphics[width=0.6\linewidth]{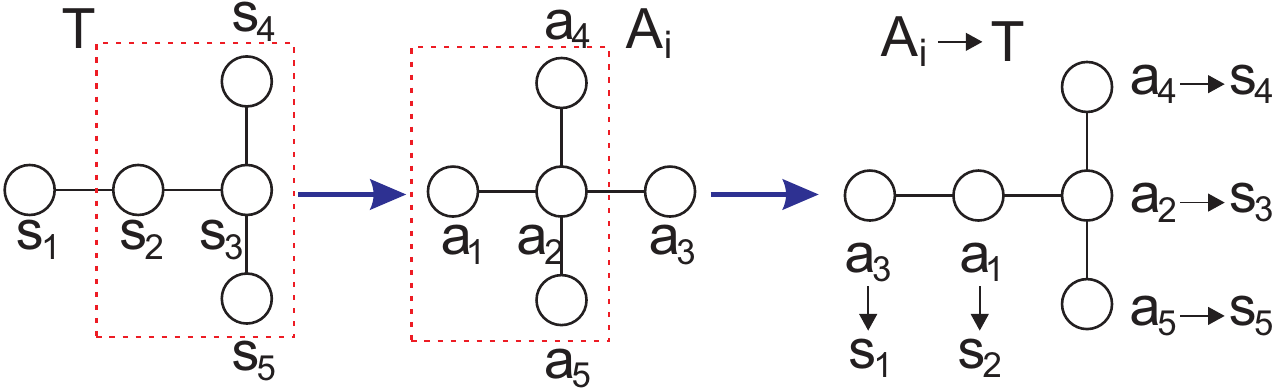}\\
(a) & (b)\\
\end{tabular}
\end{center}
\caption{(a) A scenario where the colored subgraphs of $T$ are isomorphic to $A_i$, (b) A scenario where a subgraph of $t$ is isomorphic to a subgraph of $A_i$. The red dotted box shows the maximal common subgraph between $T$ and $A_i$; the unmatched module $a_3$ is detatched from $A_i$ and allocated to spot $s_1$ by our block selection algorithm.}
\label{mcs_fault}
\end{figure}

To place $A_{curr}$ into $T$ without breaking the connections between its modules, we have to find if $T$, or a subgraph of $T$, is isomorphic to $A_{curr}.$ {\footnote{Note that if $|V_T| = |V_{A_{curr}}|$ then the problem becomes a graph isomorphism problem.}} An example of this problem is shown in Figure \ref{mcs_fault}(a) that shows all possible subgraphs of $T$ which are isomorphic to the configuration $A_i$ using different colors. This problem requires finding the isomorphic subgraphs (IS)~\cite{cordella2004sub} of $T$; it is an NP-hard problem but can be solved in polynomial-time for certain graph structures like trees~\cite{shamir1997faster}. However, if $A_{curr}$ is not isomorphic to $T$ or a subgraph of $T$, then $A_{curr}$ can not be placed into $T$ without breaking its connections and, thus, changing its shape. In such a scenario, our objective is to reduce the number of connections that are removed between $A_{curr}$'s modules. For this, we have to find the maximum number of modules in $A_{curr}$, which can be placed directly into $T$, without first disconnecting them. An example is shown in Figure \ref{mcs_fault}(b), where the red dotted boxes indicate the maximum common subgraphs of $T$ and $A_i$, which are isomorphic. This problem is an instance of the {\it maximum common subgraph (MCS) isomorphism} problem \cite{raymond2002maximum}, where, given two graphs $T$ and $A_{curr}$, the goal is to find the largest subgraph which is isomorphic both to a subgraph of $T$ and $A_{curr}$. {\footnote{If $|V_{A_{curr}}| > |V_T|$, then we find the maximum size subgraph of $T$ which is isomorphic to $A_{curr}' \subseteq A_{curr}$ and allocate the spots to matched modules, using a similar technique as in {\em blockAllocation} algorithm.}}

Our algorithm first finds subgraphs of $G_T$ that are isomorphic to $G_A$.  If there are no isomorphic subgraphs, it checks for maximal common isomorphic subgraphs. These subgraphs are stored in set $T_{sub}$ (lines $2-4$). As modules want to maximize the utility earned from the allocation, the subgraphs $t_k$ within $T_{sub}$ are ordered by utility to $A_{curr}$. The algorithm then inspects each subgraph $t_k$. If all the spots in $t_k$ are free, then $t_k$ is selected by $A_{curr}$ and $l_{curr}$ broadcasts a message to notify every module in $\mathbb{A}$ about this selection (lines $6-9$). On the other hand, if any spot $s^i \in t_k$ is already selected by a singleton $a_{block}$, $A_{curr}$ checks to see if it can evict $a_{block}$ using the {\em evict()} method. If evict is successful, $t_k$ is selected for $A_{curr}$ and the updated set of spot-selector pairs are broadcast to all modules in $\mathbb{A}$ (lines $11-15$). If eviction is not successful, it means that some modules in $A_{curr}$ could not occupy some spots in the target configuration (or its subgraph) as some other modules that did not belong to configuration $A_{curr}$ had already selected those spots. In this case, the modules of $A_{curr}$ that could not find a spot in $t_k$ will be disconnected from $A_{curr}$. Single spot selection algorithm is then used to select other spots in $t_k$ for these modules (lines $17-21$). Finally, because selection of $t_k$ by a configuration $A_{curr}$ is done by means of matching modules of $A_{curr}$ to unique spots in $t_k$, if $t_k$ is an MCS of $A_{curr}$ (i.e., $|V_{t_k}| < |V_{A_{curr}}|)$, then some of the modules in $A_{curr}$ will not be matched to any spot in $t_k$. Those unmatched modules will disconnect from $A_{curr}$, become singletons and will execute singleton module {\em spotAllocation()} algorithm, in the order of their distances from the center of $T$, to get allocated to a spot (lines $22-24$). Note that all other modules in $A_{curr}$ whose matched spots in $t_k$ were free to occupy, will occupy the matched spots while retaining their configuration. The updated set of spot-selector pairs are broadcast to all modules.

\begin{algorithm}[ht!]
{\em blockAllocation($A_{curr}, \bar{S}$)}\\
\KwIn{$\bar{S}$: Set of (spot, selector) pairs; $A_{curr}$: Set of modules connected together as a configuration and currently selecting spots.}

$T_{sub} \leftarrow$ Set of all subgraphs of $T$, which are isomorphic to $A_{curr}$.\\
\If{$T_{sub} == \{\emptyset\}$}
	{
		$T_{sub} \leftarrow$ Set of all maximum common isomorphic subgraphs of $T$ and $A_{curr}$.\\
	}
\For{each $t_k \in T_{sub}$ in descending order of utility $U_{A_i}(t_k)$}{
	\eIf{No spot in $t_{k}$ has been selected yet }
	{
		Select $t_k$;\\
		Broadcast updated set of spot-selector pairs $\bar{S}$;					
	}
	{	
		$S_{block} \leftarrow$ set of spots $\in t_k$ already selected by $\{a_{block}\} \subseteq \mathbb{A} \setminus A_{curr}$\\
		$s^i \leftarrow$ spot matched to $a_i \in A_{curr}$ but already selected by $a_{block} \in \mathbb{A} \setminus A_{curr}$\\
		\eIf{{\em evict($a_i, a_{block}$) = TRUE} for every $s^i \in S_{block}$}{
				Select $t_{k}$;\\
				Broadcast updated set of spot-selector pairs $\bar{S}$;	
		}
		{	
				\If{all $t_{k} \in T_{sub}$ has been checked}
				{
					\For{each $a_i \in A_{curr}$ where {\em evict($a_i, a_{block}$) = FALSE} and $s^i \in t_k$}{
						Disconnect $a_i$ from $A_{curr}$\\
						$A_{curr} \leftarrow A_{curr} \setminus a_i$;\\
						\em spotAllocation($a_i, \bar{S}$);\\
						Broadcast updated set of spot-selector pairs $\bar{S}$;\\
				}
		}
	}
	\If{selected $t_k$ is MCS of $A_{curr}$}{  
		\For{every $a_i \in A_{curr},$ where $s^i \not \in t_k$}{
			Disconnect $a_i$ from $A_{curr}$\\
			$A_{curr} \leftarrow A_{curr} \setminus a_i$;\\
			\em spotAllocation($a_i, \bar{S}$)} 
			Broadcast updated set of spot-selector pairs $\bar{S}$;\\
		}
		\
	}
}
\caption{Block Allocation Algorithm that a set of modules connected in configuration $A_{curr}$ uses to select a set of maximally adjacent spots in the target configuration.}
\label{algo_configuration}
\end{algorithm}
\subsection{Acting Phase}
\label{sec_acting}
After the planning phase is finished and all the spots in the target configuration have been selected by modules, the modules have to move to their respective selected spots. Note that no robot moves until all the spots are selected. If there is no proper order of robots for assuming spots, then a deadlock situation might arise. For example, in Figure \ref{fig_example}(b), if all the modules occupy their spots before module $5$ does, assuming module $5$ is a singleton, then it will be difficult for module $5$ to occupy its spot properly, unless other modules give it space for moving. But then they will have to align themselves again, which is a difficult task. To avoid this, the module which has selected the spot with highest betweenness centrality value (or, central spot), will move first and assume its position. Once it is in its proper position, it will broadcast a message to notify this to all other modules. Next the spots neighboring the center spot will be filled and so on. Techniques described in \cite{baca2014modred} can be used for locomotion of the modules.

\subsection{Analysis}
\begin{theorem}
{\em spotAllocation} and {\em blockAllocation} algorithms are complete when sufficient number of modules are available to form desired target configuration.
\end{theorem}
{\it Proof}: We prove the completeness of the algorithms by showing that there is no empty spot or hole in the target configuration when the number of modules is at least equal to the number of spots in the target configuration $T$, i.e., when $|\mathbb{A}| \geq |S|$. A hole exists in $T$ if there is a spot $s_{h}$ that is not occupied by any module. This can happen because of two conditions: $1$) No module has selected $s_{h}$, or, $2$) module $a_{h}$, which selected $s_{h}$, could not reach its spot because another module blocked the path to its selected spot by occupying a spot that was further from the center of $T$ than the selected spot. We show that these two conditions cannot arise. If $|\mathbb{A}| \ge |S|$, then because of the recursive approach in the {\em evict} method of Algorithm \ref{algo_singleton}, each module will try to select a spot in $T$, as long as there are available spots. This guarantees that condition $1$ never arises as at least one module $a_h$ will select $s_h$. Condition $2$ will never arise because, as described in Section \ref{sec_acting}, modules' priority to move is based on the betweenness centrality of their selected spots, and, spots nearer to the center of $T$ are occupied first, followed by outer ones. In other words, no module will occupy an outer spot  before its neighboring spot, that is nearer to the center of the target configuration gets occupied. Consequently, $T$ cannot have a hole. Hence proved.

\begin{theorem}
{\em spotAllocation} algorithm returns a Pareto-optimal allocation between modules and spots, i.e., any module's earned utility cannot be improved without making another module's utility worse.
\label{pareto_optimality}
\end{theorem}

{\it Proof}: Let $s_{i, k}$ denote the $k$-th highest utility spot for module $a_i$. Because each module orders the spots based on utilities, it follows that $U_{a_i}(s_{i,k}) > U_{a_i}(s_{i,k+1})$. Consider two modules $a_i$ and $a_j$ that have the highest utility for the same spot $s$ (i.e., $s_{i, 1} = s_{j, 1} = s'$, but $U_{a_i}(s') > U_{a_j}(s')$. Also, assume that $a_j$ has selected spot $s'$ first. Now, if {\em spotAllocation} allocates $a_i$ to its next best spot, $s_{i,2}$ and $a_j$ remains at $s'$, then the total utility is $U^{1} = U_{a_i}(s_{i,2}) + U_{a_j}(s')$. On the other hand, if {\em spotAllocation} method evicts $a_j$ from $s'$ and allocates it to its next best spot $s_{j,2}$ (assuming it is free), then the total utility becomes $U^{2} = U_{a_i}(s') + U_{a_j}(s_{j,2})$. From Algorithm \ref{algo_singleton}, if eviction is possible, then $U^2 > U^1$. On the other hand, if eviction does not happen, then it implies, $U^1 > U^2$. For any other allocation strategy that does not do eviction even if $U^2 > U^1$, then the total utility earned by the alternate allocation strategy is always less than the utility earned by {\em spotAllocation} algorithm. From the above equations, we can conclude that, if any two modules $a_i$ and $a_j$ have same ranking for a particular spot, $s'$, then one of the modules will be allocated to that spot and the other will be pushed to its next highest utility spot, i.e., its earned utility reduces, and no other allocation would increase their utilities as well as the overall utility. Hence the allocation strategy is Pareto-optimal.

\begin{theorem}
As $\mathcal{D}_{max}$ approaches |S|, the total utility earned by the modules ($U$) approaches the optimal utility $U^*$.
\label{social_optimality}
\end{theorem}
{\it Proof}: If there is no conflict among the modules about their best spots, i.e., each module's highest utility spot is unique, then {\em spotAllocation} algorithm allocates highest utility spots to all the modules and thus achieves the optimal utility. But if there is a conflict among modules for the same spots, then eviction method is invoked. From Algorithm \ref{algo_singleton}, we can conclude that the total utility earned by the modules increases by successively calling the evict method. For $\lim \limits_{\mathcal{D}_{max} \rightarrow |S|}$, any subsequent evictions will consequently increase the total utility. If eviction fails, then that means the total utility can not be improved any further. Thus, every time the eviction method is invoked it will increase the total utility, going towards the optimal utility. 

{\bf Note on complexity.} The {\em spotAllocation} algorithm (Algo. \ref{algo_singleton}) has a time complexity given by $O(|S|^{{\cal D}_{max}})$ where $|S|$ is the number of spots in the target configuration and ${\cal D}_{max}$ is the depth up to which the eviction of modules is allowed. In the {\em blockAllocation} algorithm (Algo. \ref{algo_configuration}), target configurations are considered to be trees and finding all possible isomorphic subtrees in the target configuration has a polynomial worst case time complexity of $O((|S||A_i|)^{D+1})$~\cite{cordella2004sub}, where $|A_i|$ and $|S|$ are the number of modules and spots in intial and target configurations respectively, and, $D$ is the maximum branch factor of either configuration ($D=3$ for our MSR).

\section{Experimental Evaluation}
{\bf Settings.} We have implemented the spot allocation algorithm on a desktop PC (Intel Core i5 -960 3.20GHz, 6GB DDR3 SDRAM). We tested instances where random number of singletons and the initial configurations with sizes between $2$ and $10$ modules need to be allocated to target configurations with between $10$ and $100$ spots. In all cases, unless otherwise mentioned, the total number of modules in the environment is equal to the total number of spots in the target configuration. Each module is modeled as a cube of size $1$ unit $\times 1$ unit $\times 1$ unit. The modules are placed at random locations within a $16$ unit $\times 16$ unit environment, their initial orientations are drawn from a uniform distribution in $\mathbb{U}[0, \pi]$, and the initial positions of singletons and leaders of the initial configurations are drawn uniformly from $\mathbb{U}[(0, 15), (0,15)]$. For all the tests, $\mathcal{D}_{max}$ has been set to $3$. Changing the value of $\mathcal{D}_{max}$ from $3$ to $10$ affected the algorithm's performance (both time and quality wise) negligibly; therefore this is not included in the results. 

Initial and target configurations were restricted to be trees based on the connections the modules in our MSR platform are capable of, although our algorithms can be applied for any other kinds of graphs as well. As there can be numerous subtrees present in the target configuration, which are isomorphic to the initial configuration and finding all possible isomorphic subtrees can take considerable time, we set an upper bound, $MAX$, on the number of isomorphic subtrees that the {\em blockAllocation} algorithm (Algo \ref{algo_configuration}) will check. $MAX$ is set to to $20$; different values of $MAX=10, 30,$ or $40$ did not change the performance of the algorithm. To get higher utility isomorphic subtrees, first the nodes in the target configuration are sorted in descending order of betweenness centrality values, because if the costs to occupy two different spots are the same, then higher betweenness centrality (spot value) indicates higher utility of the spot. For every node in the sorted list of spots, every node in current configuration $A_i$ is made the root of $A_i$ once and checked for subtree isomorphism with target configuration $T$ while making each node in $T$ the root once, for every possible tree in $A_i$. The checking of isomorphic subtrees between $A_i$ and $T$ is stopped as soon as the first $MAX$ isomorphic subtrees are found. All results are averaged over $50$ runs. 

\begin{figure}[ht!]
\begin{center}
\begin{tabular}{cc}
\hspace{-0.2in}\includegraphics[width=0.5\linewidth]{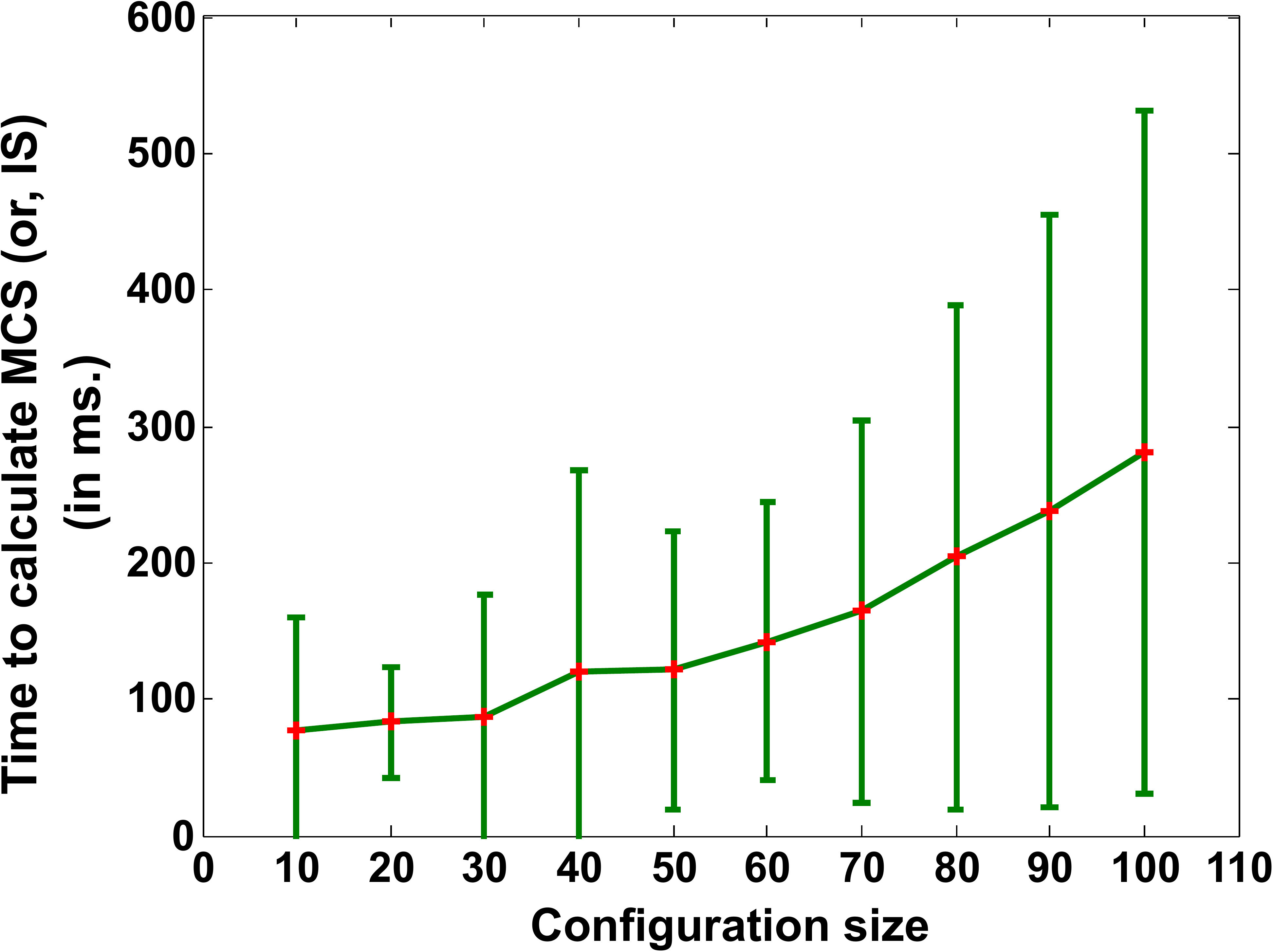}&
\includegraphics[width=0.5\linewidth]{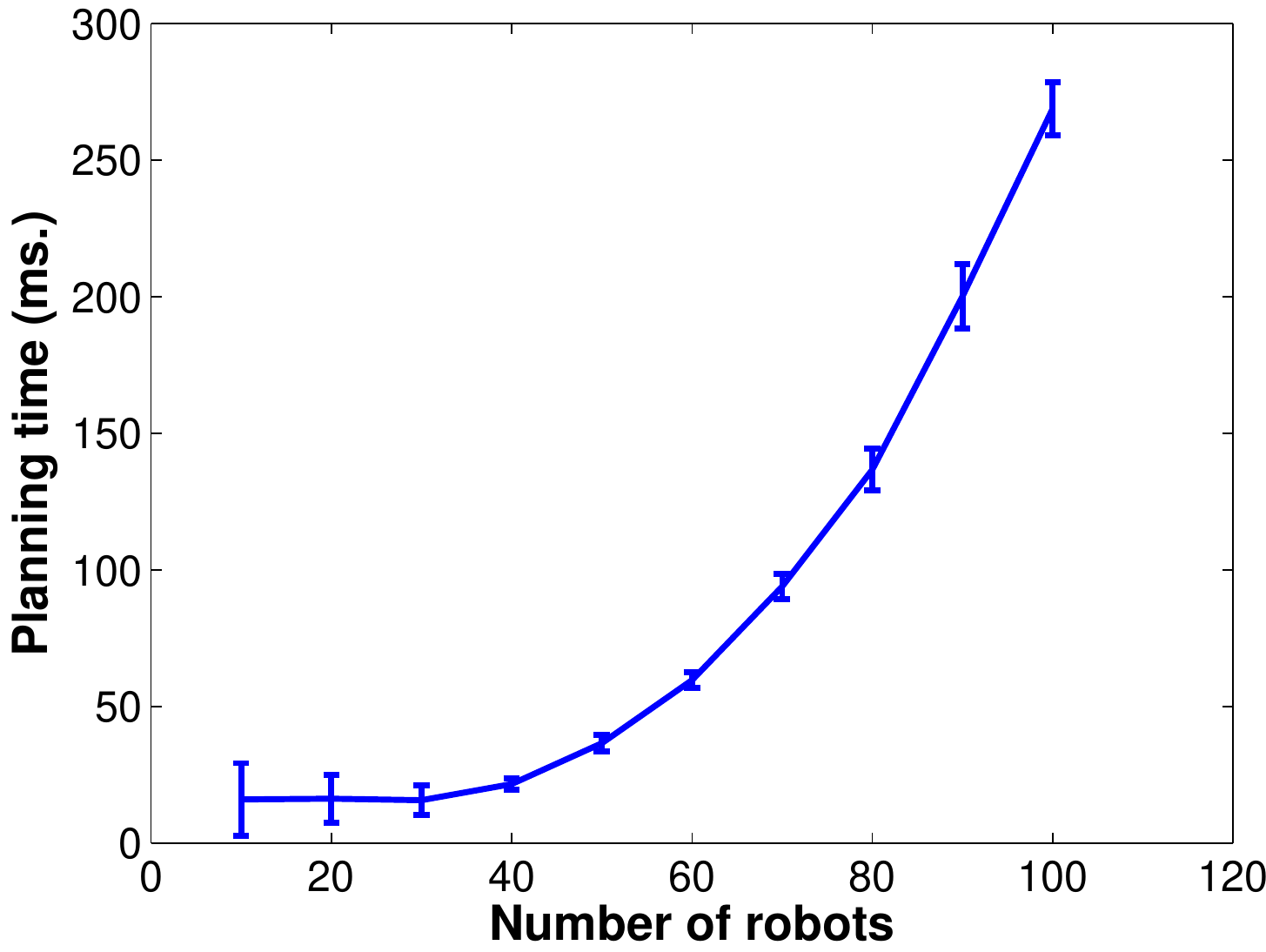}\\
(a) & (b)\\
\end{tabular}
\end{center}
\caption{(a) Time to calculate MCS or IS vs. different initial configuration sizes, (b) Total planning time for different number of modules in environment.}
\label{MCS_time}
\end{figure}

\begin{figure}[ht!]
\begin{center}
\begin{tabular}{cc}
\hspace{-0.2in}\includegraphics[scale=0.28]{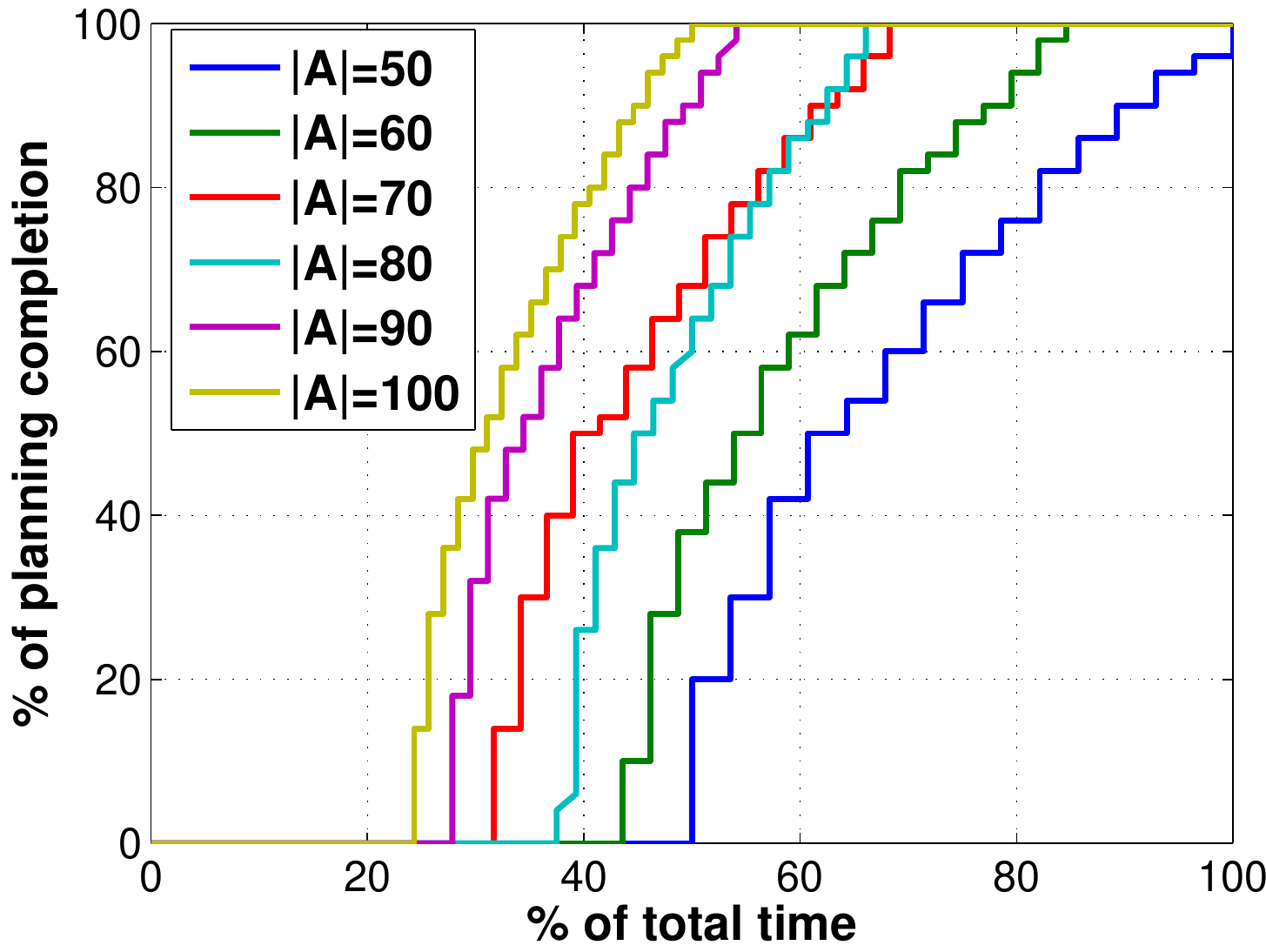}&
\includegraphics[scale=0.28]{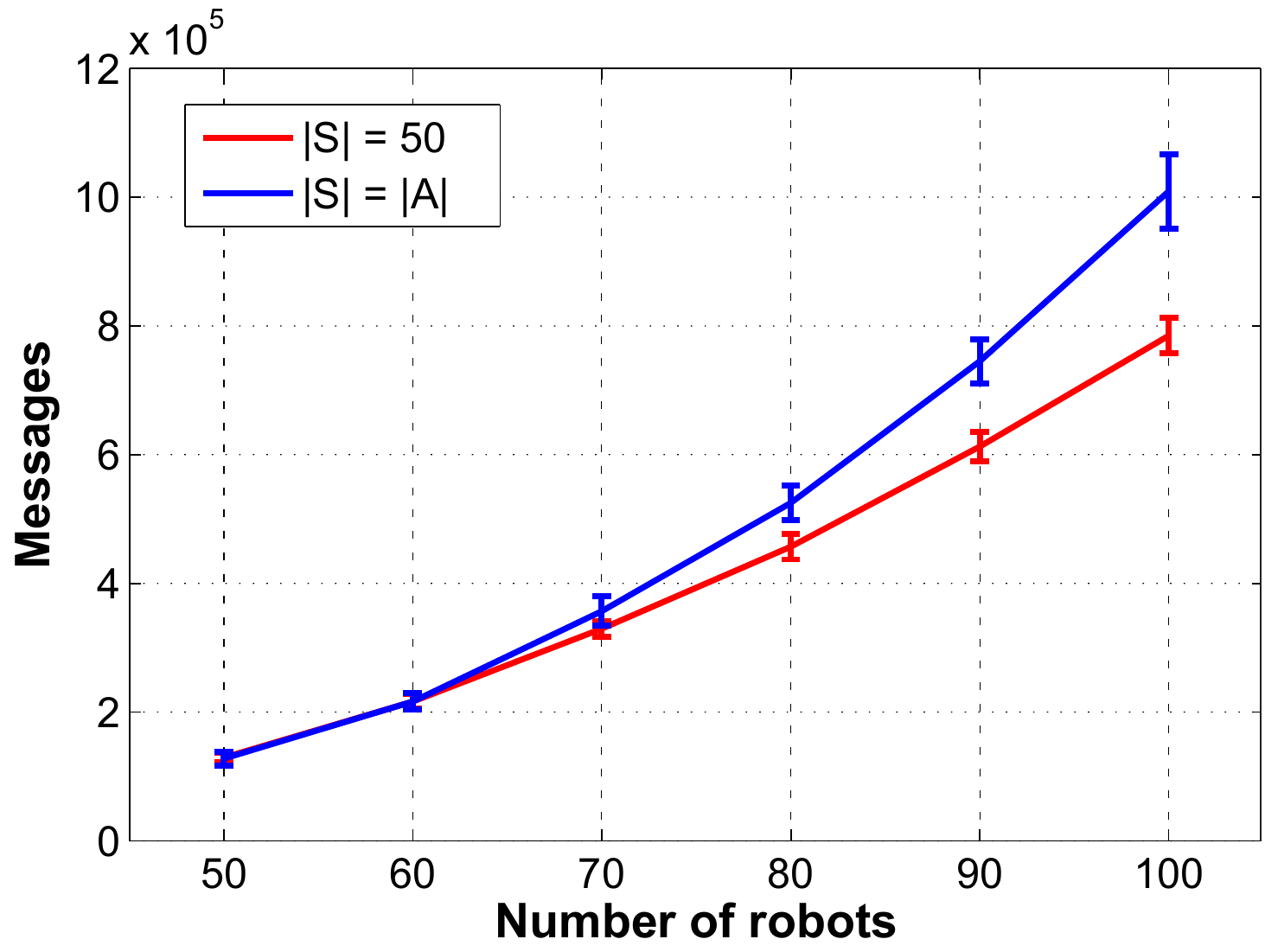}\\
(a) & (b)\\
\end{tabular}
\end{center}
\caption{(a) Change in \% of planning completion with \% of time completion, for different no. of robots and $|S|=50$; (b) Change in no. of messages for different no. of robots and different no. of spots. }
\label{result_extra}
\end{figure}

{\bf Results. Performance Analysis of Our Approach:}  First we have shown how much time it takes to find $MAX$ number of MCS (or, IS). The result is shown in Figure \ref{MCS_time}(a). The $x$-axis denotes the size of a single configuration and the $y$-axis denotes the time in milliseconds to find $MAX$ number of MCS (or, IS) of that configuration in the target configuration. For this test, total spots in the target configuration have been set to $100$. Though the run time increases with the size of the initial configuration, which can be expected because of the complexity results shown in \cite{shamir1997faster} for finding isomorphic subtrees, but still it was always well within a reasonable bound.
\begin{figure}[ht!]
\begin{center}
\begin{tabular}{cc}
\hspace{-0.2in}\includegraphics[width=0.5\linewidth]{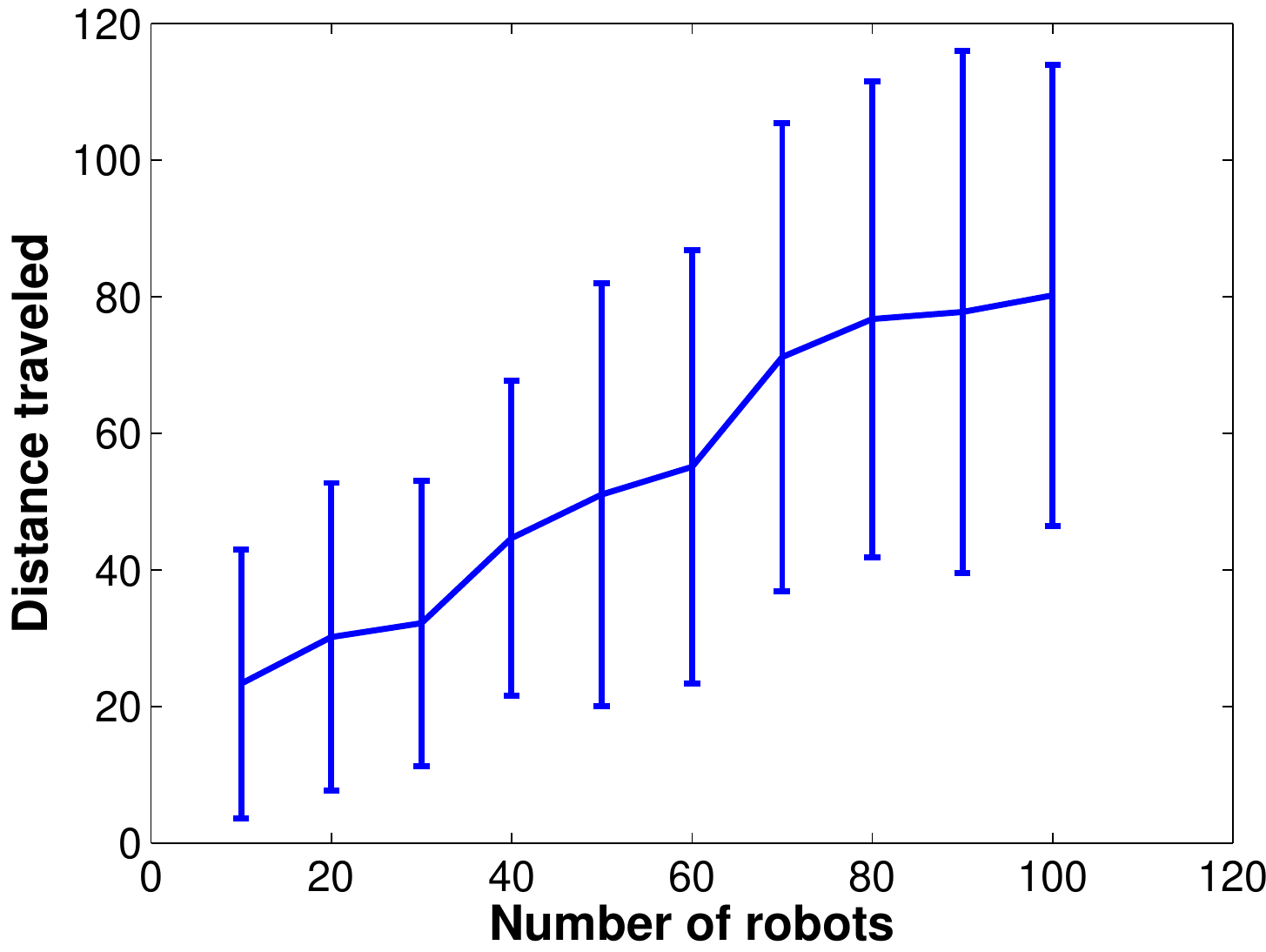}&
\includegraphics[width=0.5\linewidth]{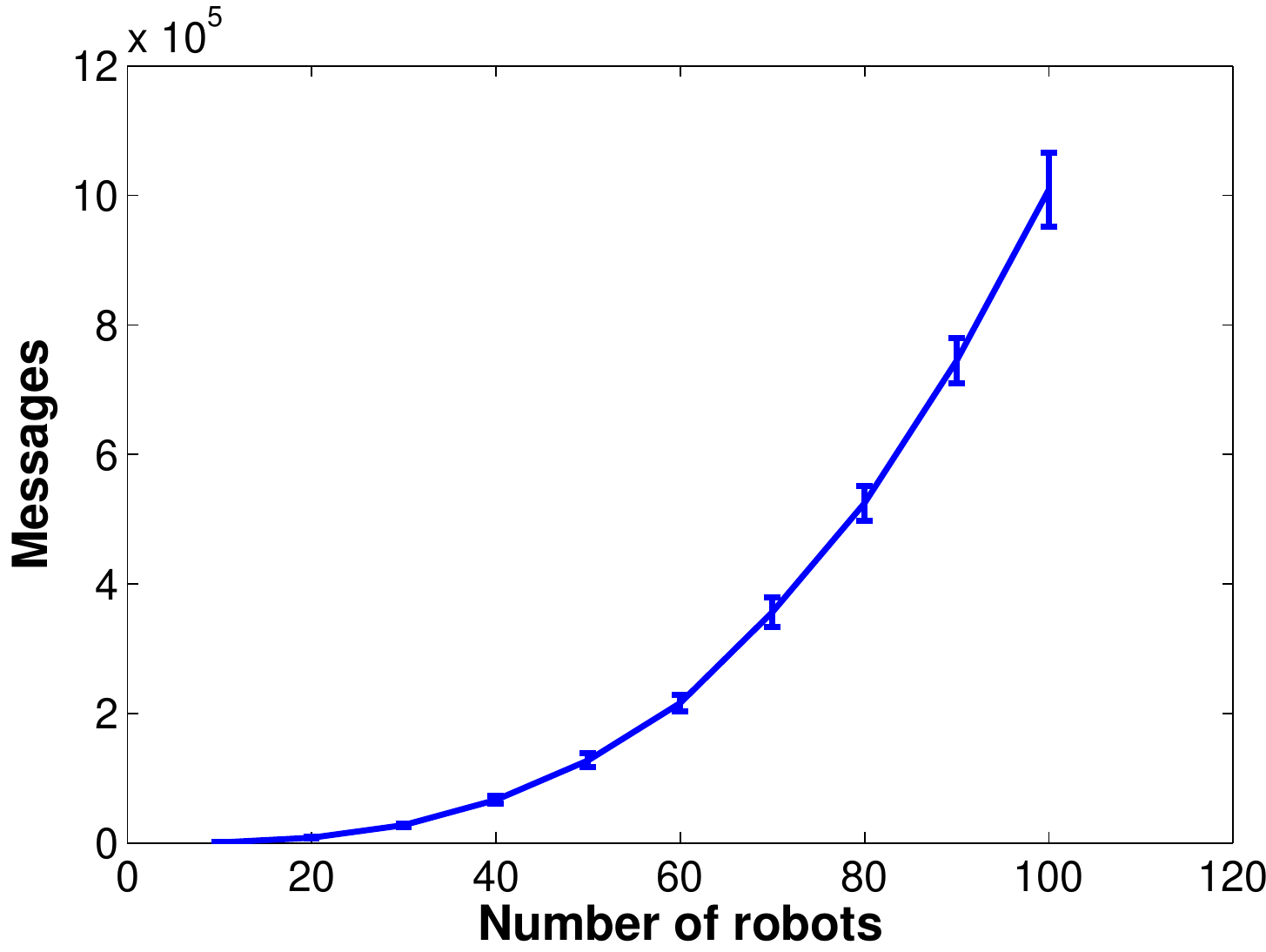}\\
(a) & (b)\\
\end{tabular}
\end{center}
\caption{{\small{(a) Distance traveled by modules to reach target configuration for different number of modules in the environment, (b) Number of messages exchanged between modules to select positions in the target configuration for different number of modules in the environment.}}}
\label{dist_msg}
\end{figure}
In the next set of experiments, we have focused on the main contribution of this paper - how to construct a modular robotic system from an initial set of singletons and configurations. Figure \ref{MCS_time}(b) shows how the planning time changes with different number of modules; the $y$-axis denotes the total planning time in milliseconds and the $x$-axis denotes the number of robots. It can be noted from this plot that though for a small set of robots, time change is almost constant, as the configuration size as well as the number of robots increases, elapsed time increases in a polynomial fashion. This elapsed time indicates only the planning phase execution time of the robots. Figure \ref{dist_msg}(a) shows how with increasing number of robots the total distance traveled by them changes. This metric is calculated by adding the distances traveled by each module from their initial positions to their respective spots in $T$. The figure shows that the total distance traveled by the robots increases linearly. We have also calculated the total number of messages passed among robots while the configuration formation process is occurring. Figure \ref{dist_msg}(b) shows how the number of total messages changes with the number of robots. As can be expected, with a higher number of modules in the environment, the number of messages increases in a polynomial fashion.
\begin{figure}[ht!]
\begin{center}
\begin{tabular}{cc}
\hspace{-0.2in}\includegraphics[width=0.5\linewidth]{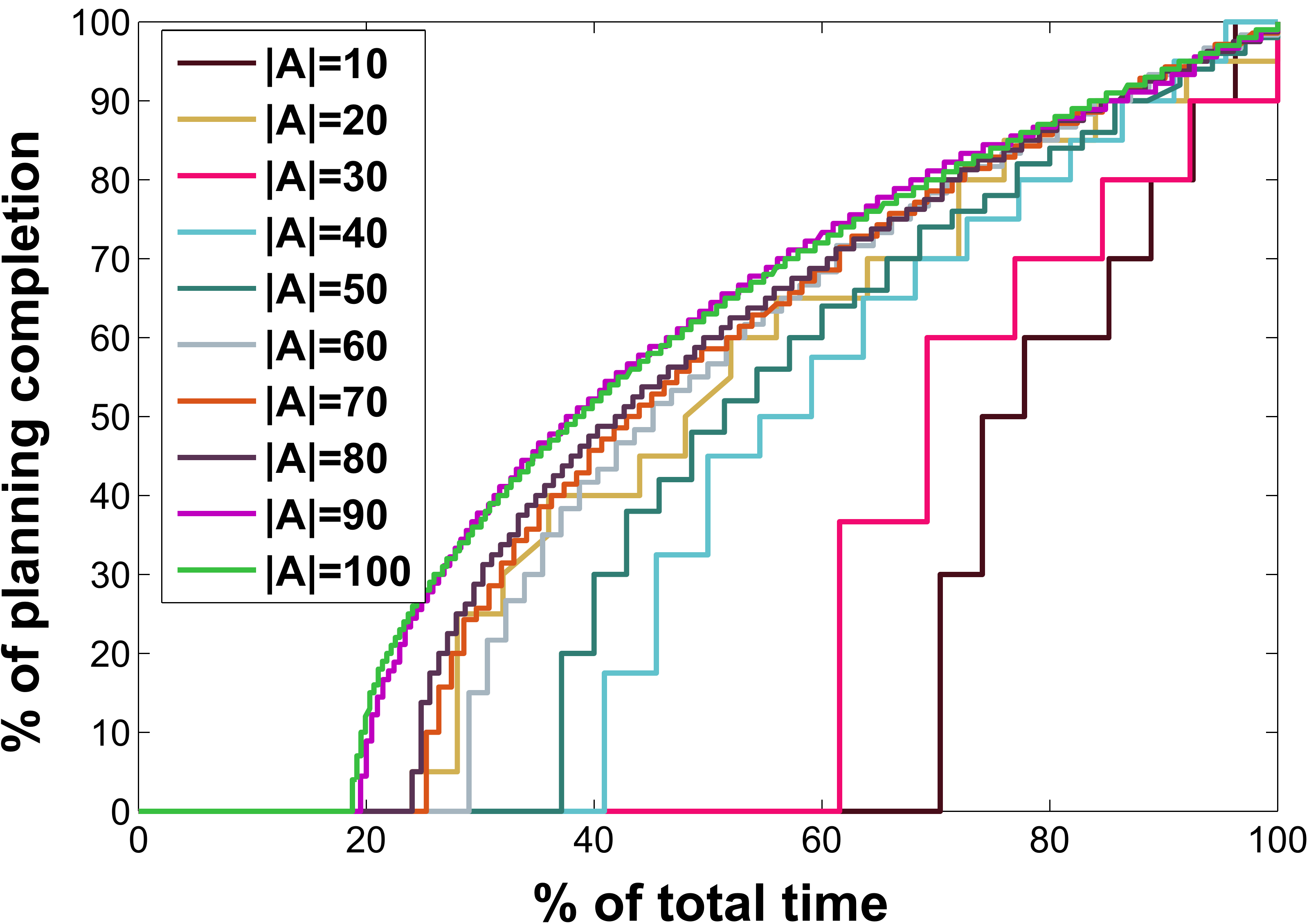}&
\includegraphics[width=0.45\linewidth]{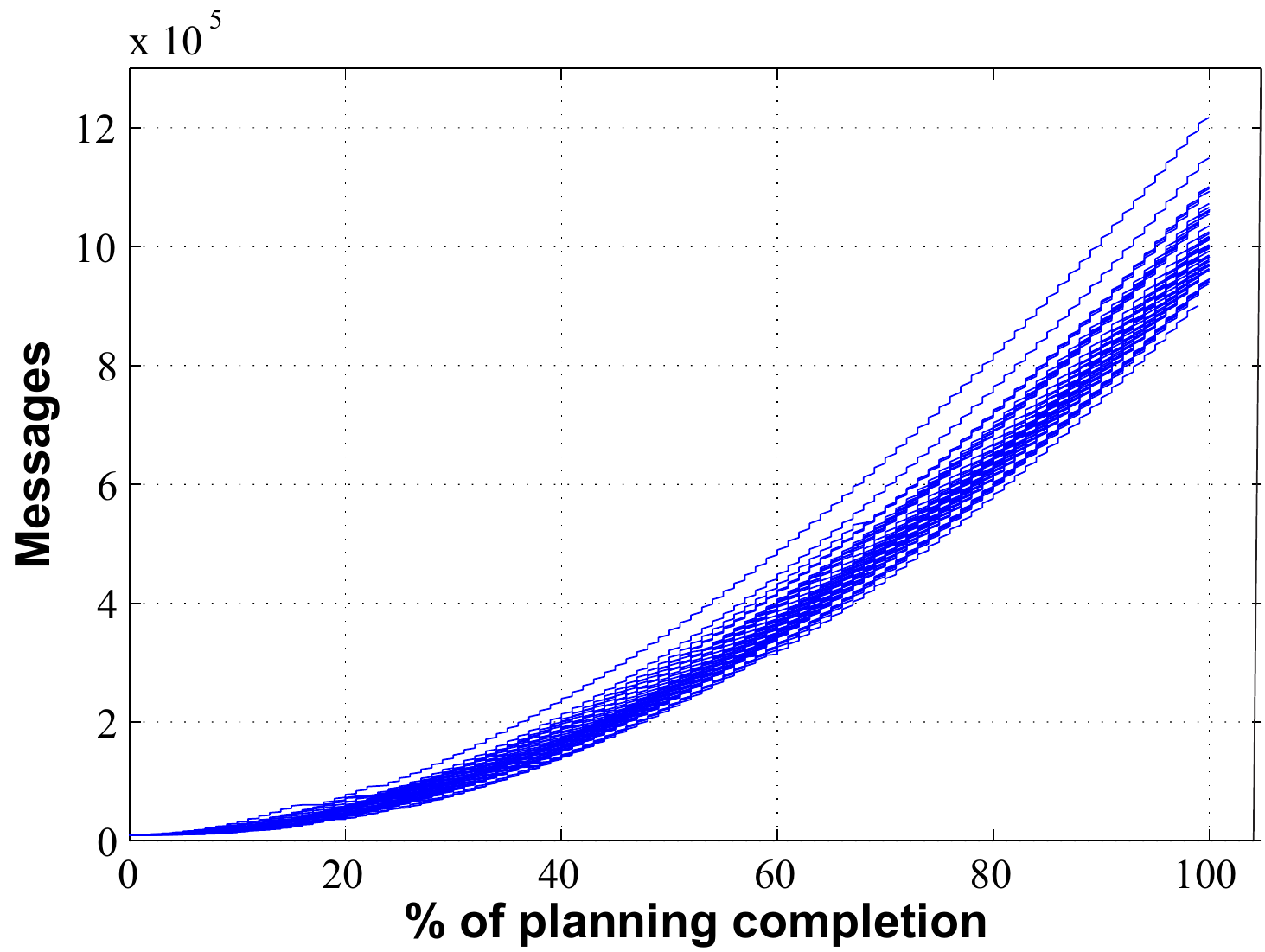}\\
(a) & (b)\\
\end{tabular}
\end{center}
\caption{{\small{(a) Change in \% of planning completion with \% of time completion, for different no. of robots; (b) Change in no. of messages at different time steps, for $100$ modules.}}}
\label{result_completion_2}
\end{figure}
Figure \ref{result_completion_2}(a) shows the planning completion rate for different number of robots between $10$ and $100$. We can see that with increasing number of robots, the completion rate increases and is more evenly distributed over time. For instance, with $|\mathbb{A}|=10$, after $70\%$ time completion, only $30\%$ of planning has been completed, whereas with $|\mathbb{A}|=100$, $30\%$ of planning gets completed only after $25\%$ of time completion. The relationship between planning phase completion and number of passed messages for $100$ modules has been shown in Figure \ref{result_completion_2}(b). All the graphs from $50$ runs have been plotted. We observe that the message count is increasing almost-linearly with completion rate.
For the next set of experiments, we have kept the number of spots, $|S|$, fixed at $50$ and we have varied the number of robots between $[50,100]$. Figure \ref{result_extra}(a) shows planning completion rate for different numbers of robots. We can see that with increasing number of robots, completion rate increases and is more evenly distributed over time. This behavior is similar to what we have seen in Figure \ref{result_completion_2}(a). Although in Figure \ref{result_completion_2}(a), for most of the robot sets, the planning phase completes almost at the end of their respective time-lines, but in the case of Figure \ref{result_extra}(a), we can notice that the planning phase finishes at different stages of their time-lines, for different numbers of robots. As an example, for $|\mathbb{A}|=100$, the planning phase almost converges at $50\%$ the of total elapsed time, whereas for $|\mathbb{A}|=50$, it takes almost $100\%$ time to converge.
Figure \ref{result_extra}(b) shows the comparison of the number of passed messages by the different numbers of robots, between the cases where $|S|=50$ and $|S|=|\mathbb{A}|$. It can be observed from this figure that with same number of robots, fewer messages are passed if there are fewer spots than robots, i.e., if $|S|<|\mathbb{A}|$. For example, with, $|\mathbb{A}|=100$ and $|S|=50$, $8\times10^5$ messages are passed, whereas with $|S|=100$ and keeping $|\mathbb{A}|$ fixed to $100$, the number of messages increases to $10\times10^5$. This result shows that the total number of messages depends on both the number of robots and spots.
\begin{table}[ht!]
\centering
\begin{tabular}{|c|c|c|}
\hline
{\small Size of All} & {\small Planning Time}  & {\small No. of Modules} \\ [0.5 ex]
{\small Initial Configurations} & {\small (ms.)}  & {\small Disconnected} \\
\hline
\hline
10 &  171.48 (avg.) & 0.12     (avg.)\\
& 15.13 (std.) & 0.32 (std.)\\
\hline
20 &  166.66 (avg.) & 4.32 (avg.)\\
& 12.88 (std.) & 3.56 (std.)\\
\hline
25 &  172.10 (avg.) & 8.76     (avg.)\\
& 11.30 (std.) & 4.85 (std.)\\
\hline
50 &  218.28 (avg.) & 29.68 (avg.)\\
& 19.57 (std.) & 5.33 (std.)\\
\hline
\end{tabular}
\caption{Planning times and the numbers of disconnected modules (average and standard deviation) in the configuration formation process, where all initial configurations have same sizes ($|S|=|\mathbb{A}|=100$).}
\label{iso_test}
\end{table}
Next we have run experiments to check how the subgraph isomorphism technique used in this work helps to reduce the number of disconnections from initial configurations. For this test, we have kept $|S|=|\mathbb{A}|=100$. Initially all modules were part of some smaller configurations and each initial configuration has the same size. We have varied the sizes of each initial configuration between $[10, 20, 25, 50]$ and thus in these cases the number of initial configurations have been varied between $[10,5,4,2]$. The planning times and number of modules required to be disconnected for these cases are shown in Table \ref{iso_test}. As can be seen, with increasing size of initial configurations, number of disconnected modules increases. This is because the probability of finding isomorphic subgraphs in $T$ decreases with increasing size of initial configurations. But the low number of disconnected modules show that it is always beneficial, in terms of number of connections detachments and re-attachments, to use our proposed approach than to break all initial configurations into singletons and then form the target configurations with them.
\begin{figure}[ht!]
\begin{center}
\begin{tabular}{cc}
\hspace{-0.2in}\includegraphics[width=0.5\linewidth]{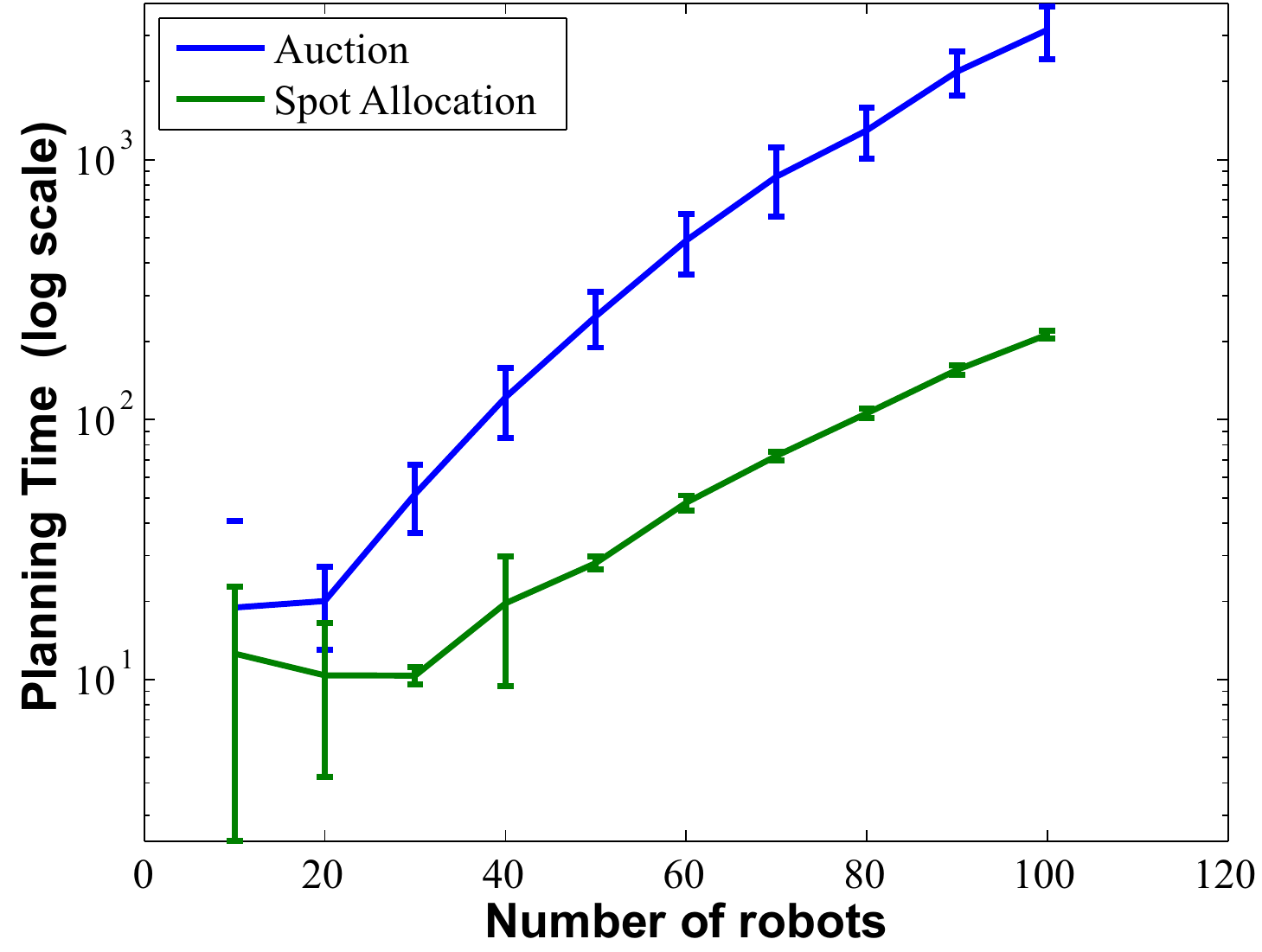}&
\includegraphics[width=0.5\linewidth]{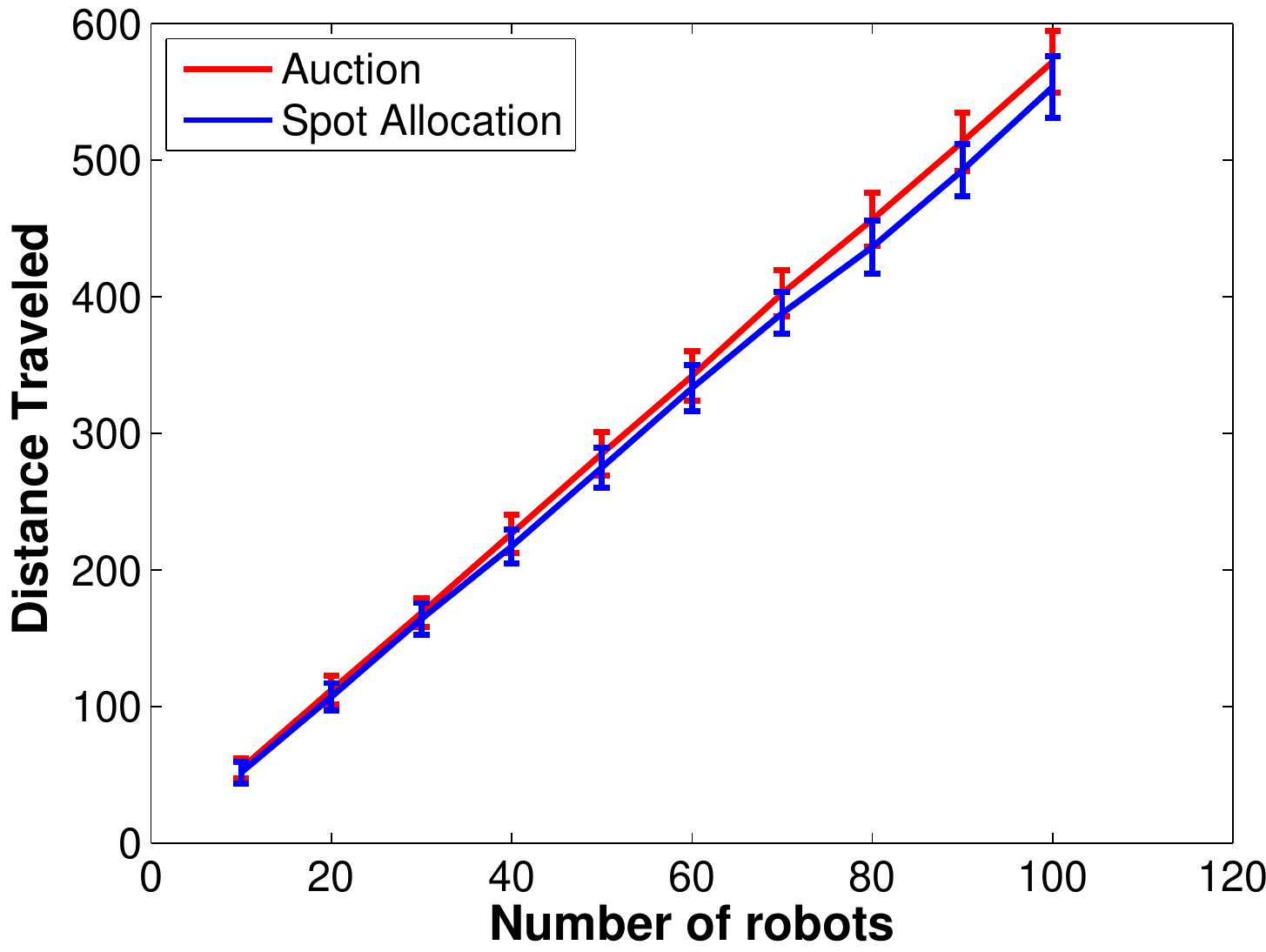}\\
(a) & (b)\\
\end{tabular}
\end{center}
\caption{(a) Log scale comparison of planning phase execution time with auction algorithm; (b) Comparison of total traveled distances with auction algorithm.}
\label{auc_comp_1}
\end{figure}

\begin{figure}[ht!]
\begin{center}
\begin{tabular}{cc}
\hspace{-0.2in}\includegraphics[width=0.5\linewidth]{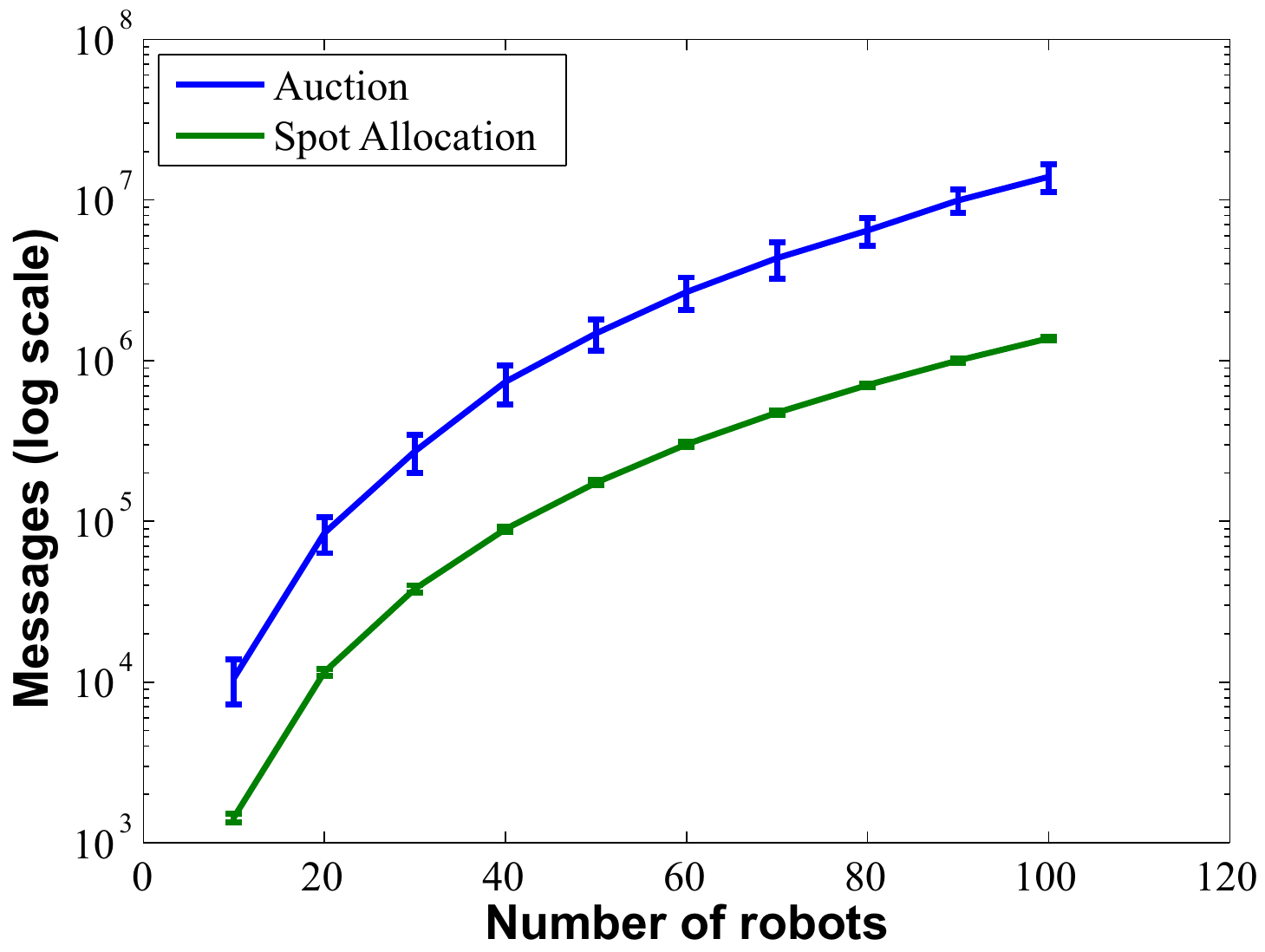}&
\includegraphics[width=0.5\linewidth]{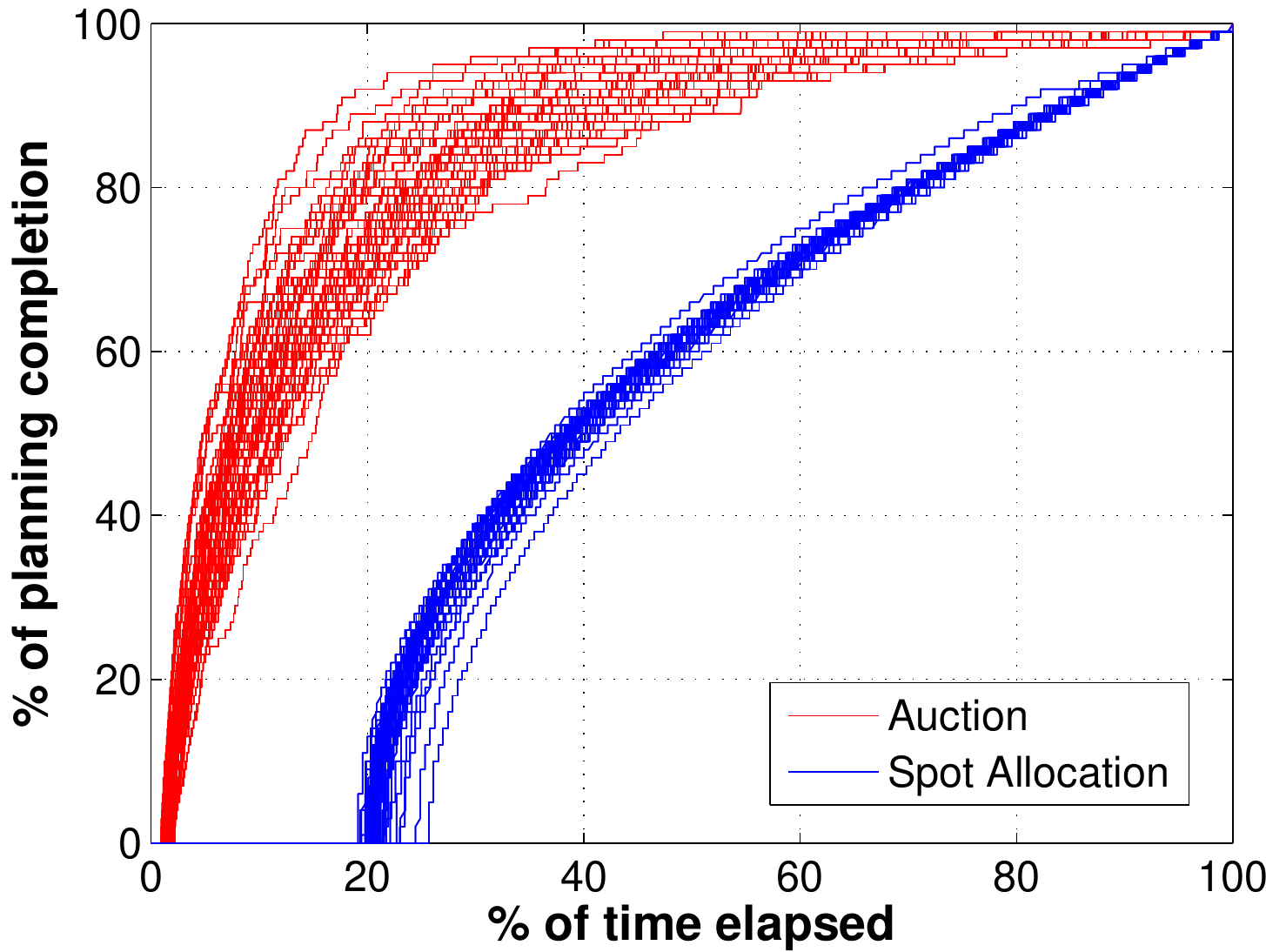}\\
(a) & (b)\\
\end{tabular}
\end{center}
\caption{{\small{(a) Log scale comparison of no. of messages with auction algorithm; (b) Change in \% of planning completion with \% of time completion and comparison with auction algorithm. $50$ lines indicate $50$ runs.}}}
\label{result_msg}
\end{figure}

{\bf Comparison with Auction-based Allocation.} We have also compared our approach for MSR configuration formation with an auction algorithm \cite{bertsekas1990auction} that finds an optimal assignment between spots and modules. Using the auction mechanism a group of modules bid for a set of spots. First the modules bid for their most preferred spots; conflict among modules for the same spot is resolved by revising bids in successive iterations. The assignment is done in a way such that the utility is maximized. The auction algorithm does not take connected configurations of modules during allocation. Therefore only for the tests which compare the performances of our algorithm against the auction algorithm, initially all the modules are considered to be singletons. A log scale comparison of planning times between spot allocation and the auction algorithms is shown in Figure \ref{auc_comp_1}(a). As can be seen from this graph, with increasing the number of robots, the difference between planning times of these two algorithms increases, i.e., our proposed algorithm's performance gets better with increased number of robots compared to the auction algorithm. Comparison of distances traveled by the modules using our algorithm and the auction algorithm is shown in Figure \ref{auc_comp_1}(b). As we can see in this plot, in most of the cases total traveled distance by the robots is the same. But with higher numbers of robots, using the proposed spot allocation algorithm robots travel less distance than by using the auction algorithm. Thus the spot allocation algorithm assigns the spots to the modules in very nominal time, keeping the cost for movement almost the same (or less in some cases), compared to the auction algorithm. A log scale comparison of number of the messages generated, by the spot allocation and auction algorithms, is shown in Figure \ref{result_msg}(a). This figure indicates that the spot allocation algorithm generates fewer messages than the auction algorithm, which helps to reduce the communication overhead. Figure \ref{result_msg}(b) compares the completion rates of planning phases of the auction and spot allocation algorithms - the $x$-axis denotes the percentage of total time elapsed. This result indicates that completion rate of the auction algorithm is higher, even though  the auction algorithm takes longer than the spot allocation algorithm.

\begin{figure*}[thb!]
\begin{center}
\begin{tabular}{c|c}
\hline
\includegraphics[width=2.2in]{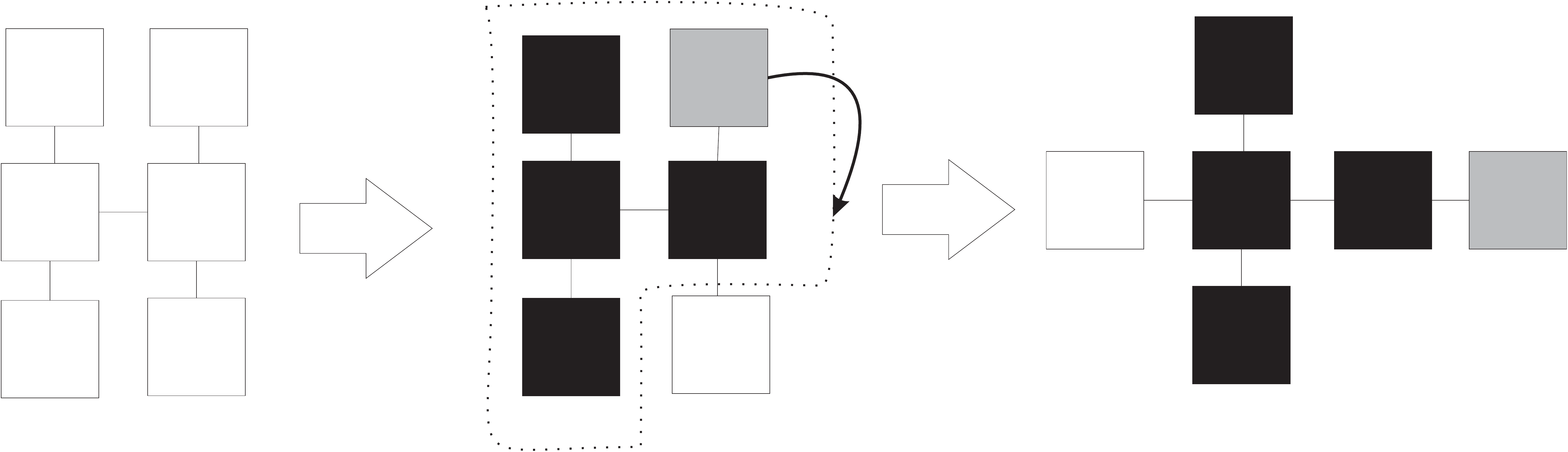}&
\includegraphics[width=2.2in]{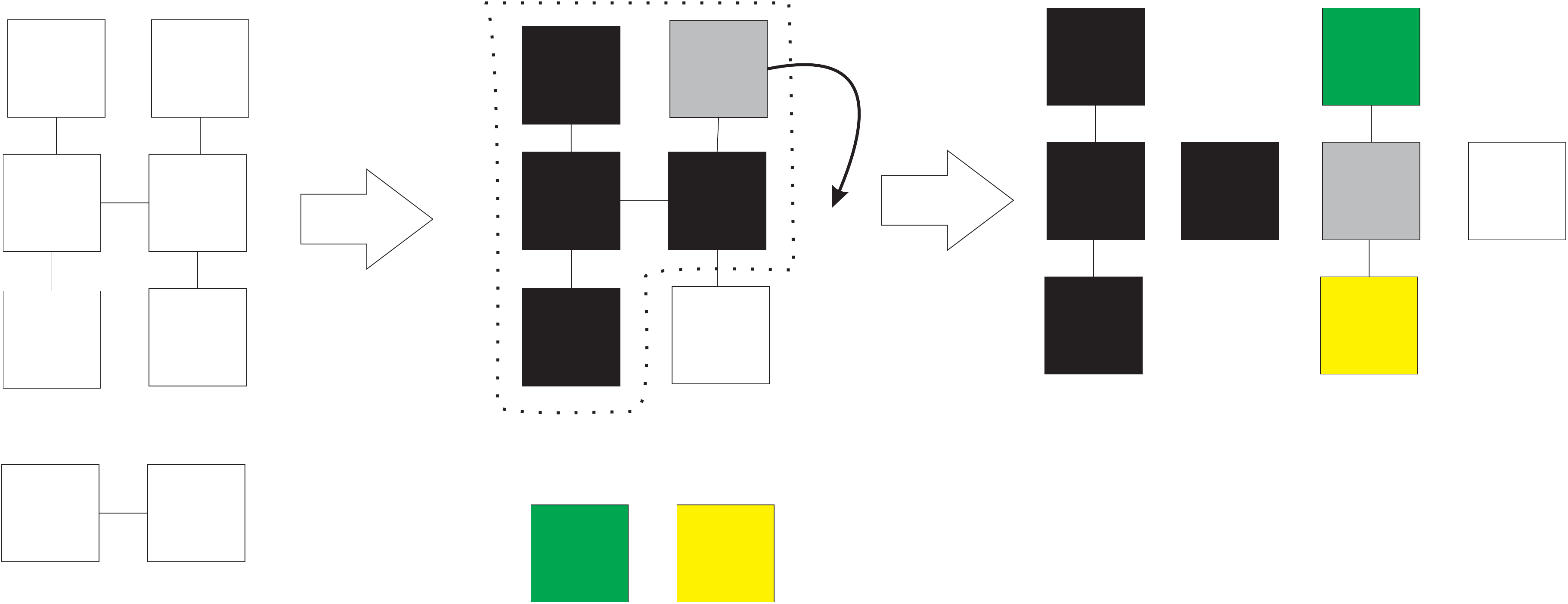}\\
Case $1$: Planning Time: $110$ ms., No. of disconnections: $1$ & Case $2$: Planning Time: $113$ ms., No. of disconnections: $2$\\
\hline
\includegraphics[width=2.2in]{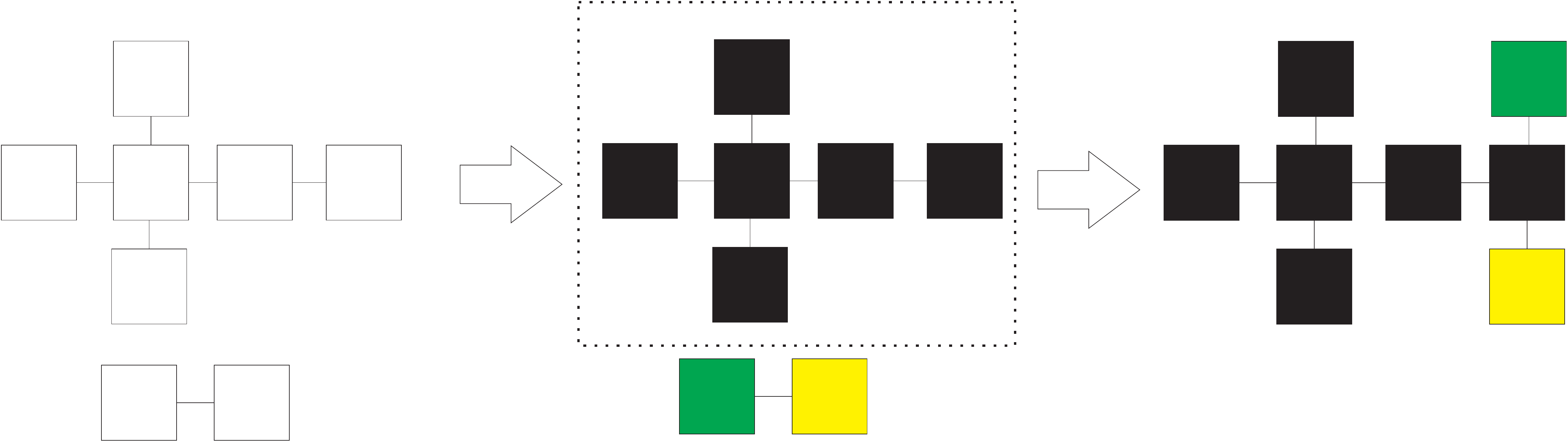}&
\includegraphics[width=2.2in]{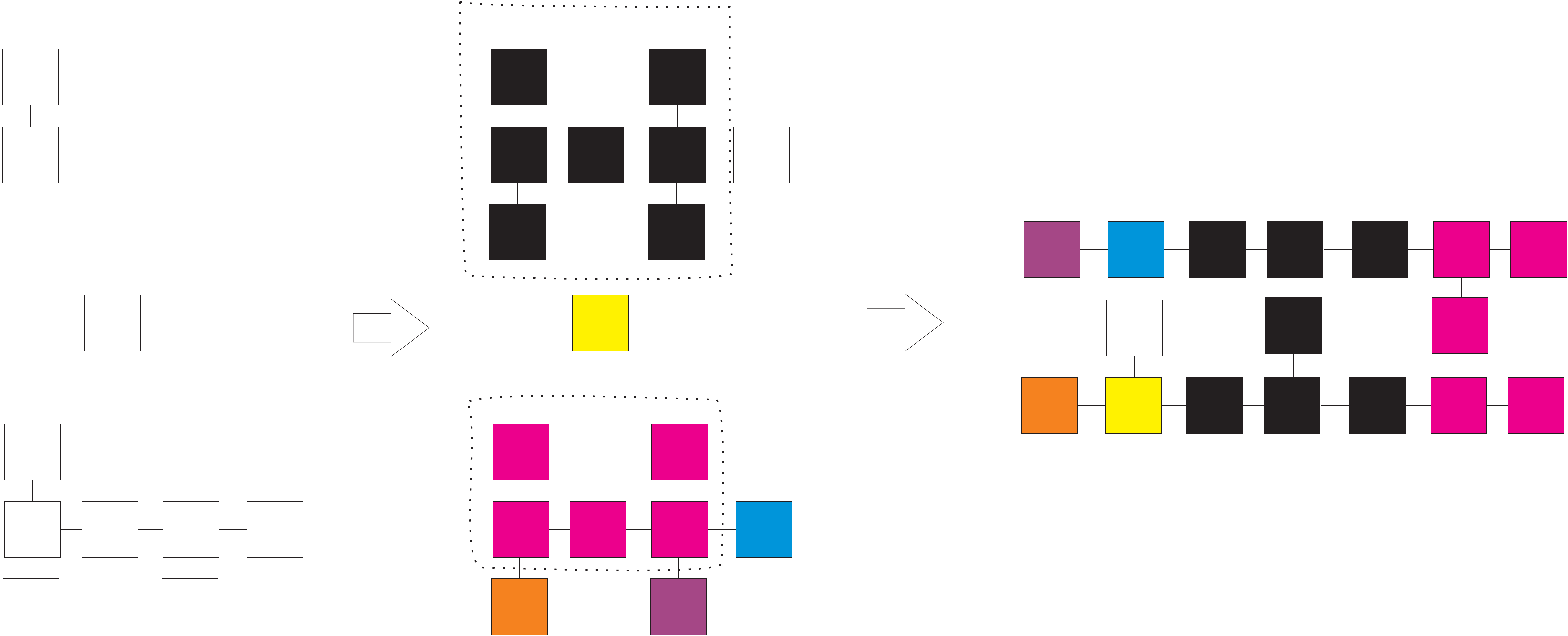}\\
Case $3$: Planning Time: $111$ ms., No. of disconnections: $1$ & Case $4$: Planning Time: $170$ ms., No. of disconnections: $4$\\
\hline
\includegraphics[width=2.2in]{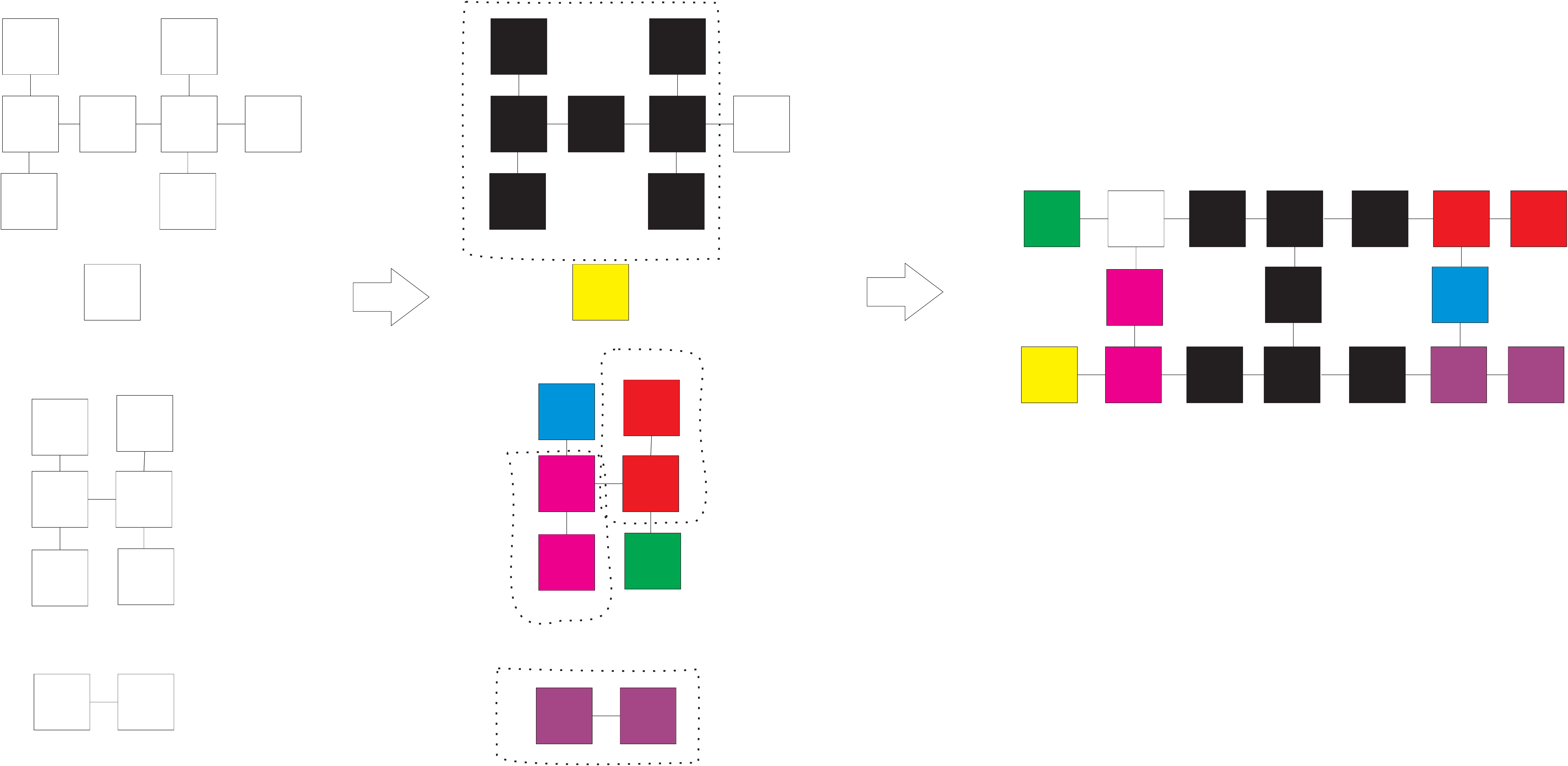}&
\includegraphics[width=2.2in]{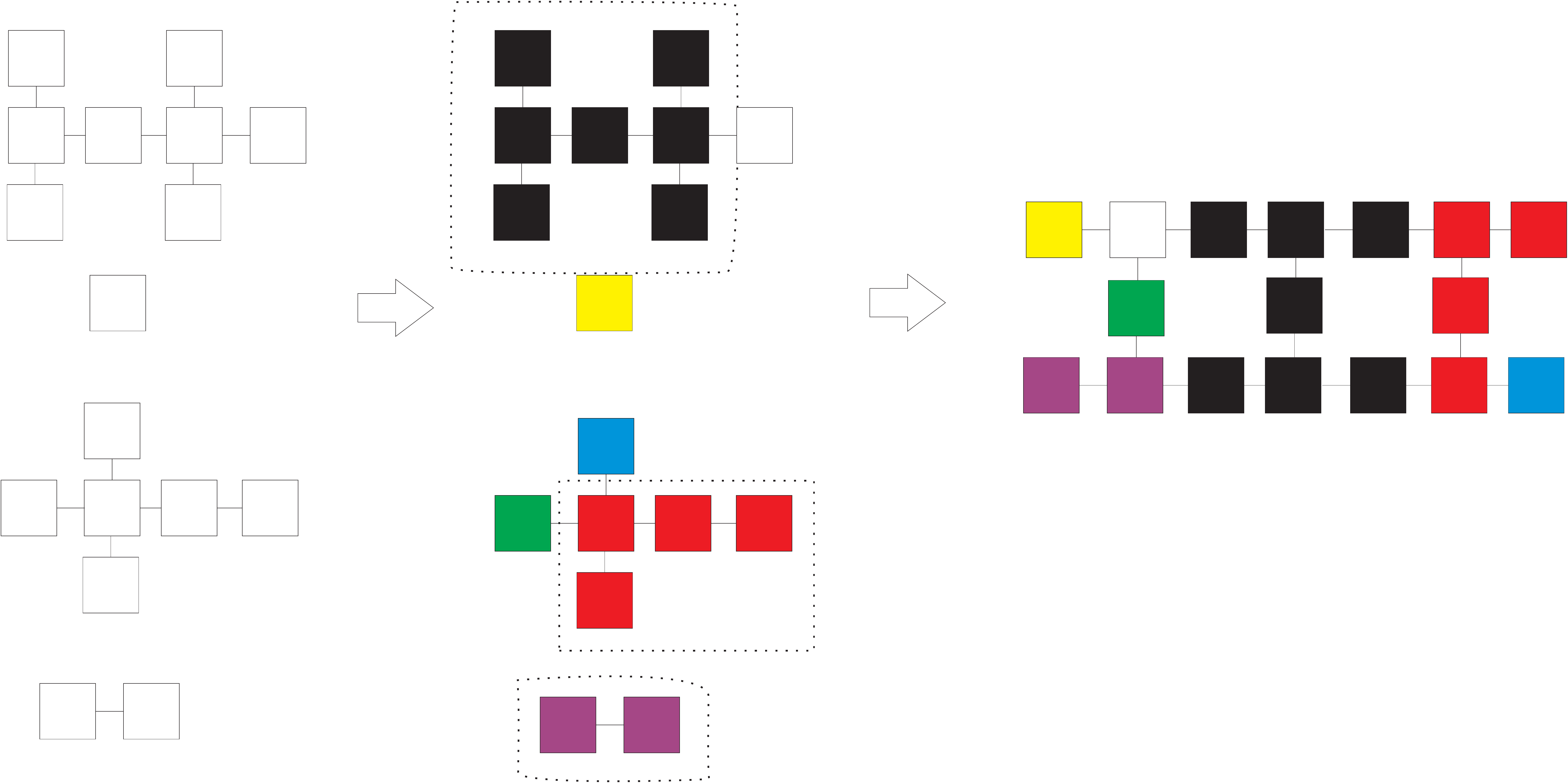}\\
Case $5$: Planning Time: $182$ ms., No. of disconnections: $3$ & Case $6$: Planning Time: $173$ ms., No. of disconnections: $3$\\
\hline
\includegraphics[width=2.2in]{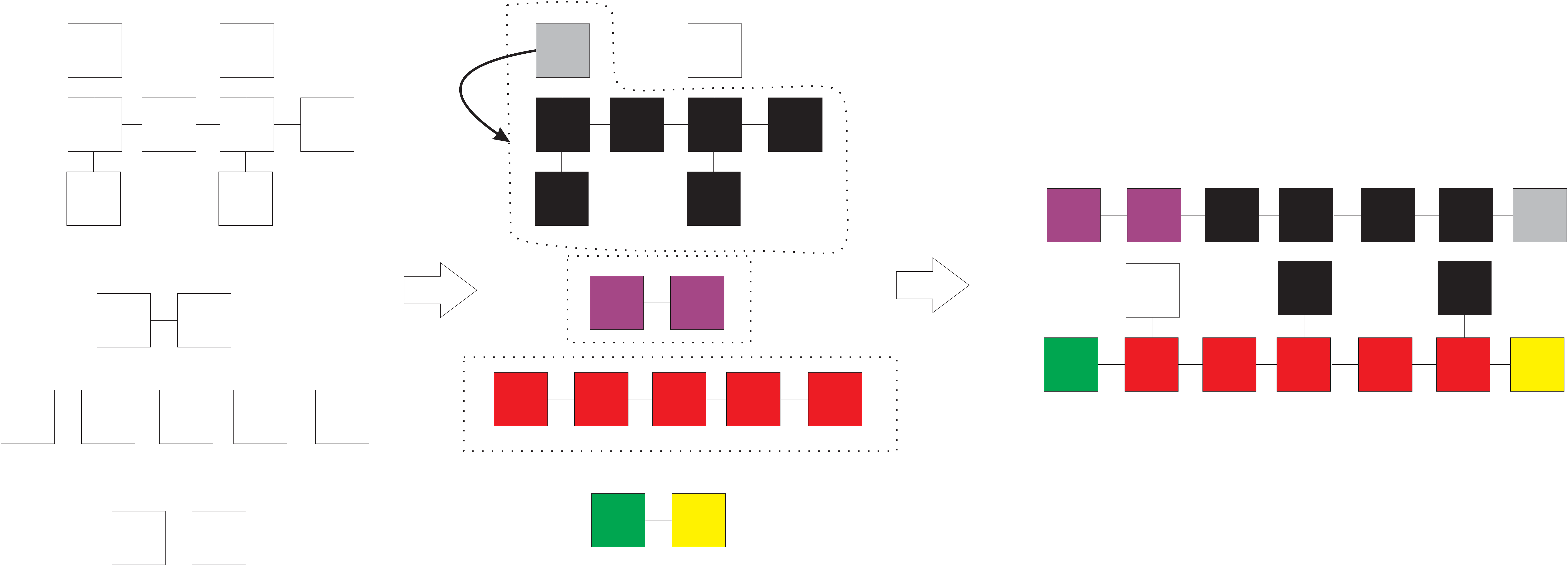}&
\includegraphics[width=2.2in]{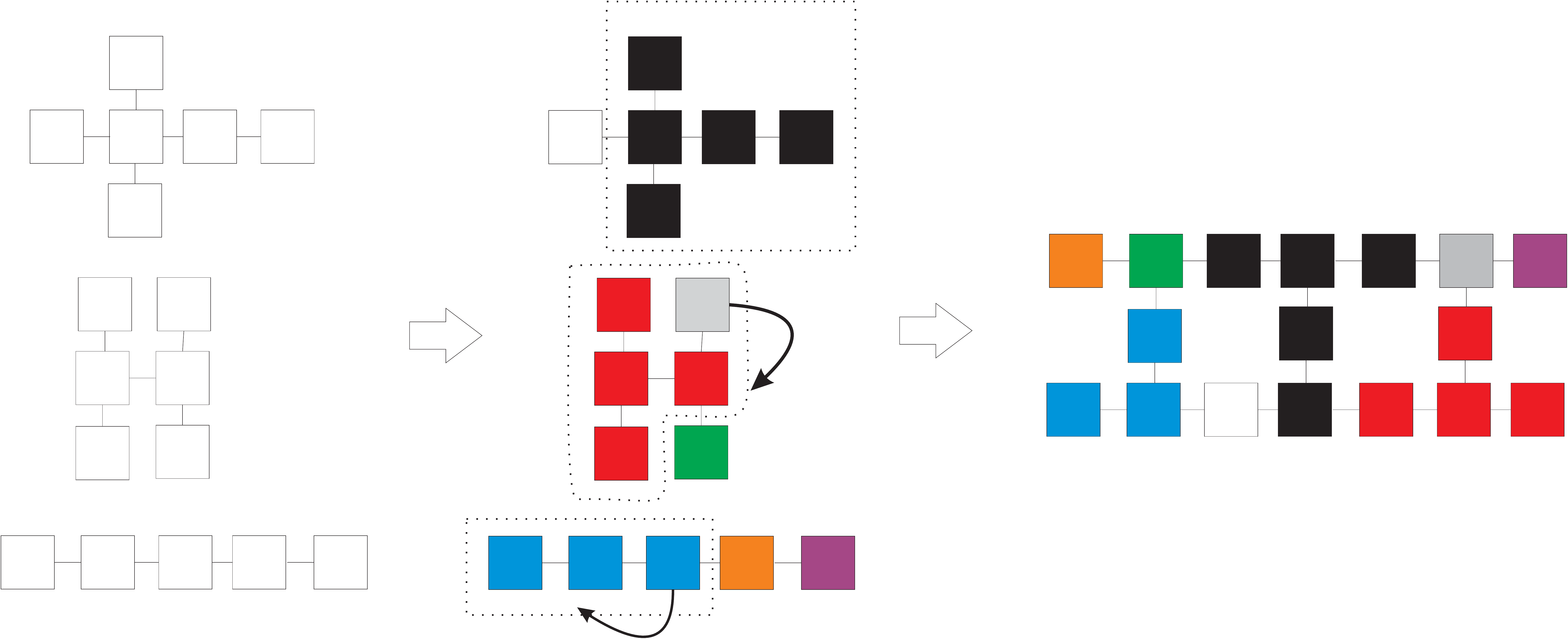}\\
Case $7$: Planning Time: $190$ ms., No. of disconnections: $2$ & Case $8$: Planning Time: $184$ ms., No. of disconnections: $4$\\
\hline
\end{tabular}
\end{center}
\caption{{\small{Cases showing configuration formation procedure along with corresponding planning times and number of disconnections required. Leftmost figure in each case shows the initial configurations and singletons, middle figure shows the MCS (or, IS) found (marked by dotted boxes) by executing our algorithms, rightmost figure shows the final formed target configuration with modules selecting spots (shown in a color-coded fashion).}}}
\label{cases}
\end{figure*}

\vspace{-0.1in}

\subsection{Case Studies}
In this section, we have shown $8$ specific cases of the configuration formation process that are shown in Figure \ref{cases}. Each of the initial and target configurations used for this set of experiments have been shown to be feasible and stable in~\cite{hossain2014kinematics}. Squares represent the modules and the links between two squares denote the connection between those two modules. For each case illustrated, the left-most diagram shows the initial configurations and/or singletons, the middle diagram shows the detected MCS (or, IS) and the diagram on the right shows the final formed configuration. The modules are color-coded to show the final allocations. MCS (or, IS) are shown with dotted boxes. Grey-colored modules represent the modules that remain connected to the same neighboring module between initial and target configurations, but only change the connector through which they are connected. Although this operation requires one undocking and one redocking operation, it consumes less energy than if the module were to be connected to a non-neighbor module. The planning time and number of disconnections for each case are provided alongside each configuration formation case in Figure \ref{cases}. We can see that each of the test cases requires less than $200$ milliseconds of planning time. Target configurations are also formed with relatively low number link disconnections (maximum being $4$).

\vspace{-0.1in}
\section{Conclusion and Future Work}
In this paper, we have proposed a novel spot allocation algorithms for configuration formation in MSRs. In the future, we are planning to add uncertainty in module's movements and an MDP-based technique for the planning phase of our algorithm. We also plan to consider the problem where there could be multiple target configurations and each module needs to decide which spot in which target configuration they should select. We are also planning to investigate algorithms that optimize the battery spent in getting to the target configuration and investigate the trade-off between aborting target configurations and saving battery. Finally we are working on implementing this algorithm on physical MSR hardware.

{\small 
\bibliographystyle{abbrv}
\bibliography{refs_RD}
}
\end{document}